\documentclass[11pt]{article}

\PassOptionsToPackage{table}{xcolor}
\usepackage[final]{acl}

\usepackage{times}
\usepackage{latexsym}

\usepackage[T1]{fontenc}
\usepackage[utf8]{inputenc}

\usepackage{microtype}

\usepackage{inconsolata}

\usepackage{graphicx}
\usepackage{tabularx}
\usepackage{fvextra}
\usepackage{adjustbox}
\usepackage{array} 
\usepackage{tcolorbox}
\usepackage{booktabs}
\usepackage{multirow}
\usepackage{amsmath}
\usepackage{xcolor}
\usepackage{wrapfig}
\usepackage{pifont} 
\usepackage{hyperref}
\usepackage{amssymb}         %
\usepackage{caption} %
\tcbuselibrary{skins, breakable} %
\newcommand{\cmark}{\ding{51}}
\newcommand{\xmark}{\ding{55}}

\title{Probe to Act: Elevating Browser-Use Agent via Active Visual Probing}

\author{
  \textbf{Keliang Li\textsuperscript{1,2}},
  \textbf{Heng Wang\textsuperscript{3}},
  \textbf{Chen Hu\textsuperscript{3,\textdagger}},
  \textbf{Daxin Jiang\textsuperscript{3}}, \\
  \textbf{Hong Chang\textsuperscript{1,2,\textdagger}},
  \textbf{Shiguang Shan\textsuperscript{1,2}} \\
  {\normalfont \textsuperscript{1}Institute of Computing Technology, Chinese Academy of Sciences} \\
  {\normalfont \textsuperscript{2}University of Chinese Academy of Sciences} \\
  {\normalfont \textsuperscript{3}StepFun} \\
  {\small\normalfont \textsuperscript{\textdagger}Corresponding authors: 
  \href{mailto:hatcher@stepfun.com}{hatcher@stepfun.com},
  \href{mailto:chonghong@ict.ac.cn}{chonghong@ict.ac.cn}}
}

\begin{document}
\maketitle
\begin{abstract}
Browser-use agents require seamless alignment between structured web metadata %
and visual information, %
while preserving relevant %
context across long interactions. Existing interfaces often rely on either screenshot-level action prediction or static Set-of-Marks overlays, leaving the model to resolve dense DOM-pixel alignment before every operation. 
We introduce \textbf{Probe to Act (P2A)}, an active probing framework for the browser-agent loop that moves this alignment into decision time. 
P2A addresses an asymmetric bridge between symbolic DOM hypotheses and screenshot layout by rendering on-demand symbolic DOM structure back into pixels. 
Before committing a state-changing browser operation, the agent can issue lightweight probes to translate DOM handles into pixel evidence, map screen regions back to DOM candidates, register visual-only targets, and commit verified notes. 
These interleaved processes naturally produce evidence-based memory: only probed, acted-on, or explicitly committed observations are kept across steps, preserving only decision-critical evidence in long-horizon contexts.
P2A can be used as a prompting strategy for proprietary models under the standard DOM+SoM interface, and can be distilled into open-weight models through cold-start synthesis and self-bootstrapped SFT. 
Across three browser-use benchmarks, P2A shows clear gains on task success rate for both proprietary and fine-tuned models; on VisualWebArena, for example, it improves Gemini-3-Pro from 54.1\% to 61.2\% and Qwen3-VL-8B from 24.6\% to 32.9\%, while matching the costly full-observation history ($\sim$3$\times$) at only $\sim$1.2$\times$ the peak retained input context of action-only history.
\end{abstract}

\begin{figure*}[!t]
    \centering
    \includegraphics[width=\linewidth]{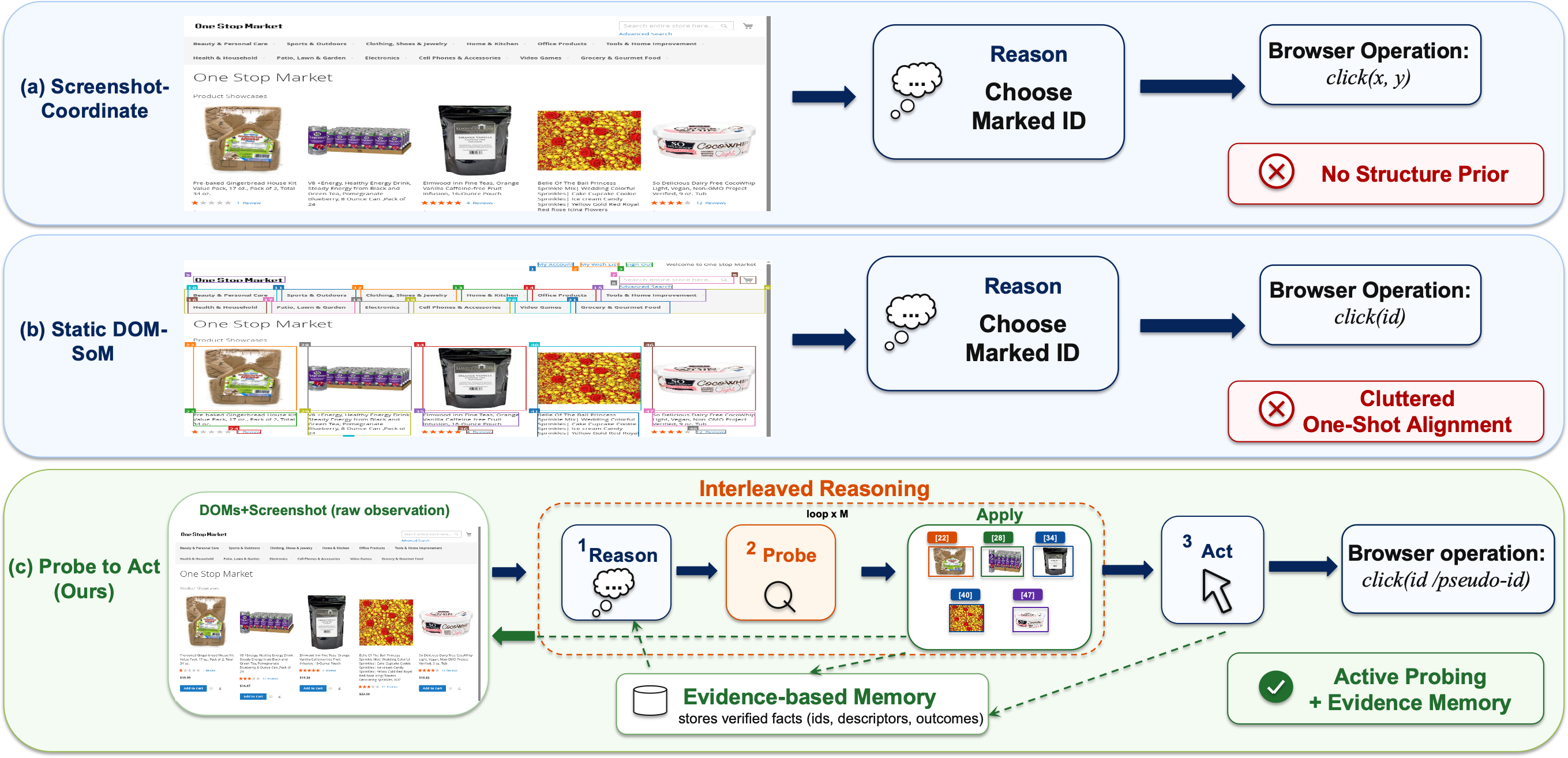}
    \caption{Passive grounding vs.\ active visual probing. (a) Screenshot-coordinate policies infer actions from raw pixels, discarding page structure; (b) static DOM-SoM overlays all marks at once, cluttering dense pages and forcing one-shot alignment; (c) our \textbf{Probe to Act} interleaves \emph{reason}--\emph{probe} loops that locally ground DOM cues on demand and retain only verified evidence in memory before acting.}
    \label{fig:comparison}
    \vspace{-1.2em}
\end{figure*}

\section{Introduction}
\label{sec:intro}
Multimodal Large Language Models (MLLMs) \cite{bai2025qwen3,bai2025qwen2,chen2024internvl2.5,hurst2024gpt} have catalyzed autonomous browser-use agents that execute natural-language goals in realistic web environments~\cite{zhou2023webarena,koh2024visualwebarena,qi2024webrl}. 
Yet web navigation remains fundamentally challenging because webpages can contain hundreds of tightly packed elements whose semantics and states must be read from the visual interface\cite{koh2024visualwebarena,deng2023mind2web}, while successful execution often requires long-horizon interaction tracking\cite{sun2025scaling,song2026compress}.
Critically, agents could observe the page through two complementary views---structured metadata (e.g., DOM/accessibility trees) and pixels (screenshots). 
A robust agent should exploit both: metadata provides efficient semantic priors and symbolic handles for interaction, whereas pixels provide the spatial layout and visual-only cues that metadata may miss.

A core bottleneck is that most agents resolve DOM--pixel alignment \emph{statically}. 
Coordinate-based policies act on pixels alone~\cite{liu2025scalecua,qin2025ui}, losing structural priors; DOM-centric policies act on text trees~\cite{zhou2023webarena,qi2024webrl}, which are efficient when clean but fragile under noisy, incomplete webpages where key targets may be missing or visually defined.
Hybrid baselines attempt to align the two views by overlaying a full-page Set-of-Marks (SoM)~\cite{yang2023set,koh2024visualwebarena} but force a static global alignment that clutters and occludes the view, and offer no way to verify ambiguous candidates or recover targets absent from the DOM \cite{zhang2025prune4web,song2026compress}. Figure \ref{fig:comparison} contrasts these three observation paradigms.
Progress in high-level frameworks such as planning, skill learning, and context compression~\cite{yang2024agentoccam,akkil2024agent,gu2024your,zhang2025prune4web,song2026compress} has improved reasoning and efficiency, but largely treats perception as a passive input. This leaves a key perspective still open: \emph{reasoning-time, on-demand DOM--pixel alignment} for effective browser use.

To this end, we introduce \textbf{Probe to Act (P2A)}, an active visual probing framework that adapts visual prompting from a static overlay into an active, internal reasoning process. The key challenge is asymmetric: mapping pixels to text is native to multi-modal agents, but verifying a symbolic element by rendering it back onto the screenshot is the under-exploited direction. 
As in Figure \ref{fig:comparison}, P2A therefore enables the agent to \textbf{select what to align, when to align it, and in which direction} with probing primitives for visualization: before any browser operation, the agent actively renders candidates as pixel evidence, inspects screen regions against DOM structure, or registers visually salient regions not contained in the DOM.

Crucially, P2A also provides a principled mechanism for long-horizon memory. 
Instead of discarding historical observations or retaining expensive full screenshots and DOM trees \cite{yang2024agentoccam,song2026compress}, we introduce an \emph{evidence-based memory} strategy that stores only \emph{verified evidence} when reasoning---elements or regions that were inspected, acted upon, or explicitly anchored during probing. 
This yields a compact, decision-faithful interaction history that preserves operation-critical cues while substantially reducing context cost.

We operationalize P2A through two complementary paths: training-free prompting for proprietary models and cold-start fine-tuning plus self-bootstrapped distillation for open-weight models. Empirically, probing leads to consistent improvement on various benchmarks, e.g. lifting Gemini-3-Pro from 54.1\% to 61.2\% and Qwen3-VL-8B from 24.6\% to 32.9\% on VisualWebArena (VWA). Its evidence-based memory also matches full-observation retention at $\sim$1.2$\times$ the peak retained input context of action-only history, compared with $\sim$3$\times$ for full histories. Our contributions are summarized as below:
\begin{itemize}
    \item \textbf{An active probing interface for asymmetric DOM--pixel bridge:} P2A aligns DOM structure and screenshots within reason--probe loops, making the agent verify cross-view evidence during decision-making instead of relying on static one-shot alignment.
    \item \textbf{Evidence-based memory:} P2A retains only probed, acted-on, or committed evidence across steps, matching full-observation history ($\sim$3$\times$) at $\sim$1.2$\times$ the peak retained input context of action-only history. 
    \item \textbf{Empirical gains and analysis:} P2A obtains consistent improvement across browser navigation benchmarks, %
    showing that learning probe trajectories matters beyond only final-operation supervision.
\end{itemize}
In summary, P2A demonstrates that equipping browser-use agents with on-demand cross-modality probing and probe-driven memory yields consistent improvements across proprietary and open-weight models, at minimal additional context cost.

\section{Related Work}
\label{sec:related_work}
\subsection{Browser-Use Agent}
Web navigation agents convert natural-language instructions into browser operations within realistic websites, where observation modalities define a key design axis. Text-centric agents act over DOM or accessibility trees derived from raw HTML~\cite{zhou2023webarena,qi2024webrl,yang2024agentoccam}, which are efficient on clean pages but brittle when targets are visually defined or absent from the parse. Screenshot-native agents instead emit actions directly from pixels~\cite{hong2024cogagent,qin2025ui,liu2025scalecua,gu2025ui}, preserving visual fidelity at the cost of structural priors. Hybrid agents bridge the two by a static full-page Set-of-Marks (SoM)~\cite{yang2023set} that ties DOM indices to on-screen boxes~\cite{koh2024visualwebarena,zheng2024seeact,he2024webvoyager}, while orthogonal work improves planning, tool use, exploration, and data pipelines~\cite{prabhu2025walt,akkil2024agent,gu2024your,yu2024exact,he2025openwebvoyager,pahuja2025explorer,xiao2025uigenie}. P2A keeps the hybrid interface but removes its static-overlay assumption: DOM--pixel alignment is performed on demand inside the decision loop, so candidates can be verified and off-DOM targets recovered before acting.

\subsection{Visual Reasoning and Active Perception}
Recent visual-reasoning work treats perception as an active test-time process, allowing models to crop, zoom, or draw while reasoning~\cite{hu2024sketchpad,menon2024whiteboard,wu2025reinforcing,zhang2025thyme,zhu2025active,zheng2025deepeyes,li2026humanpcr,Li_2026_CVPR}. GUI-Eyes and related GUI methods learn screenshot crop or zoom for fine-grained visual grounding~\cite{xu2025visual,jiang2025zoomclick,chen2026guieyes}, while SGV verifies and refines trajectory-level behavior~\cite{andrade2025let}. P2A instead acquires DOM--pixel evidence within each step of long-horizon web navigation, so these approaches operate at complementary layers.
\subsection{Context Management for Long-Horizon Agents}
Long-horizon web agents reduce context cost by selecting observations before acting~\cite{zhang2025prune4web,ouyang2026focusui,song2026compress}, or by compressing past trajectories into summaries, embeddings, or cross-session memory~\cite{sun2025scaling,wu2025autoscaling,liu2025webcoach,yang2024agentoccam}. Prune4Web performs one-way, screenshot-conditioned DOM pruning, whereas ColorBrowserAgent combines progressive summarization with human-in-the-loop knowledge adaptation~\cite{wang2026colorbrowseragent}. P2A instead ties within-task memory to the decision loop: only probed, acted-upon, or explicitly committed evidence is retained, while long-horizon RL remains an orthogonal direction~\cite{wei2025webagentr1,ye2025mobile}.
\begin{figure*}[htb]
    \centering
    \includegraphics[width=0.95\linewidth]{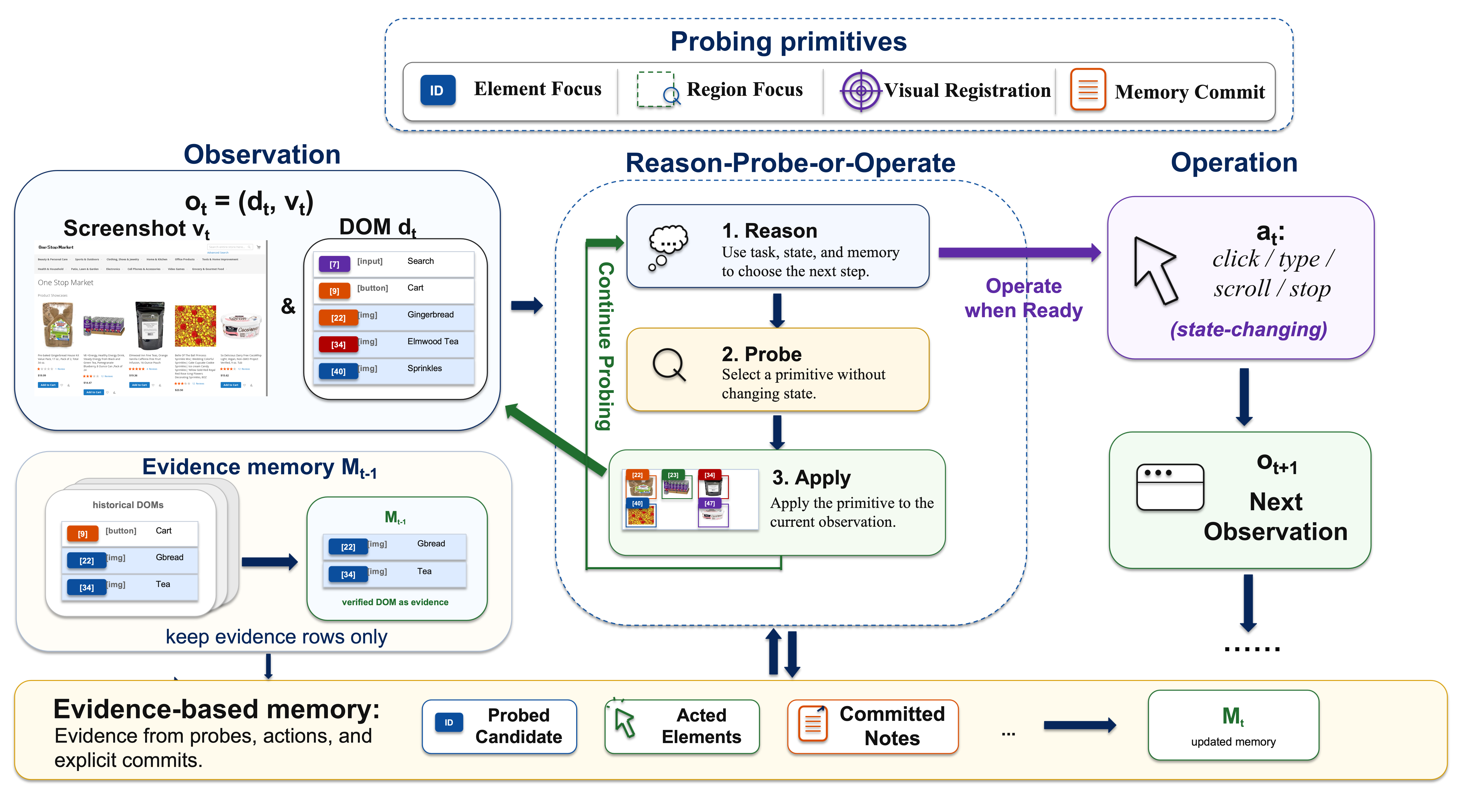}
    \caption{Overview of Probe to Act. At each step the agent observes the screenshot $v_t$ and DOM $d_t$, then enters a state-preserving \emph{Reason--Probe--Operate} loop: it reasons, selects a probing primitive, and applies it to the current observation, repeating until ready to emit a state-changing browser operation $a_t$. Verified rows are distilled into a sparse evidence memory $M_t$ that persists across steps.}
    \label{fig:overview}
    \vspace{-1.2em}
\end{figure*}
\section{Method}
\label{sec:method}
\subsection{Preliminaries}
Following prior works \cite{koh2024visualwebarena,andrade2025let}, we formulate the browser-use task as a sequential decision-making process over macroscopic environment steps. At step $t$, the agent observes $o_t = (d_t, v_t)$, where $d_t$ is the structured DOM text and $v_t$ is the page screenshot. Given the system prompt $s$, user instruction $q$, the history $H_{t-1}$, and the current observation $o_t$, the agent policy $\pi_\theta$ produces a browser operation $a_t \in \mathcal{A}_{\text{browser}}$ (click, type, scroll, etc.) along with a textual rationale $r_t$:
$$(r_t, a_t) \sim \pi_\theta(\cdot \mid s, q, H_{t-1}, o_t). $$
Standard baselines overlay a static Set-of-Marks (SoM) \cite{yang2023set} on $v_t$, rendering bounding boxes and index labels for all interactable DOM elements. This provides a fixed global alignment between $d_t$ and $v_t$ but clutters dense pages and cannot verify ambiguous candidates or recover targets absent from the DOM.\\
Due to the high token cost of retaining full observations, most prior works restrict the history $H_{t-1}$ to past actions and rationales within a sliding window, discarding all prior screenshots and DOM trees.

\label{sec:setup}
\subsection{The Probe-to-Act Loop}
\label{sec:overview}
Figure~\ref{fig:overview} illustrates our overall pipeline. The core idea of P2A is to let the agent \textit{verify cross views}. Unlike static SoM, which fixes DOM--pixel alignment once before the step, P2A defers alignment into the decision loop, resolving it locally and only where the reasoning demands. We posit that an explicit cross-modal bridge on the two observation modalities---structured DOM text $d_t$ and the visual screenshot $v_t$---benefits the agent, which must actively reconcile its beliefs from both views before committing to an action. P2A realizes this bridge through a set of probing primitives $\mathcal{A}_{\text{probe}}$ (Sec. \ref{sec:primitives}) that locally augment the current observation on demand---rendering bounding boxes to visually ground DOM entries, extracting focused crops for high-resolution inspection, or annotating textual notes onto the screenshot---all without changing the page state. By interleaving such probes with textual reasoning, the agent progressively aligns its semantic and visual understanding of the current page, and only acts when it has gathered sufficient cross-modal evidence.
Moreover, the set of elements inspected during probing naturally yields a compact, decision-faithful memory that persists across page transitions (Sec. \ref{sec:memory}).

\paragraph{Interleaved Probing and Reasoning.} Probing primitives $\mathcal{A}_{\text{probe}}$ and browser operations $\mathcal{A}_{\text{browser}}$ reside in a unified action space. At each reasoning turn, the agent freely chooses to either probe for more evidence or commit a browser action. A probe action $p_{t,k} \in \mathcal{A}_{\text{probe}}$ (the $k$-th probe action in operation step $t$) deterministically refines the internal observation $o_{t,k} = \mathrm{apply}(p_{t,k},\; o_{t,k-1})$, interleaved with textual reasoning segment $r_{t,k}$. The resulting chain $\tau_t = (r_{t,0},\, p_{t,1},\, r_{t,1},\, \ldots,\, p_{t,K},\, r_{t,K},\, a_t)$ terminates when the agent emits a browser operation $a_t \in \mathcal{A}_{\text{browser}}$. The key asymmetry is that probes are state-preserving (modifying only the agent's view) while $a_t$ is state-changing (navigating the page, rendering prior overlays obsolete).
\paragraph{Step-Local Context.} Within a step, all probe overlays and reasoning accumulated so far are retained---the agent sees the full history of its verification on the current page. Across steps, stale overlays are discarded; only the DOM metadata of probed or acted-upon elements persists as evidence memory (Sec. \ref{sec:memory}), and inter-step history keeps only the final rationale $r_{t,K}$ and $a_t$. This design has two immediate benefits: probes make actions more deliberate (verifying before clicking avoids costly mistakes such as deleting wrong items), and the set of elements touched during probing provides a principled signal for what should be remembered.

\subsection{Probing Primitives}
\label{sec:primitives}
\begin{figure}[!t]
    \centering
    \includegraphics[width=0.93\linewidth]{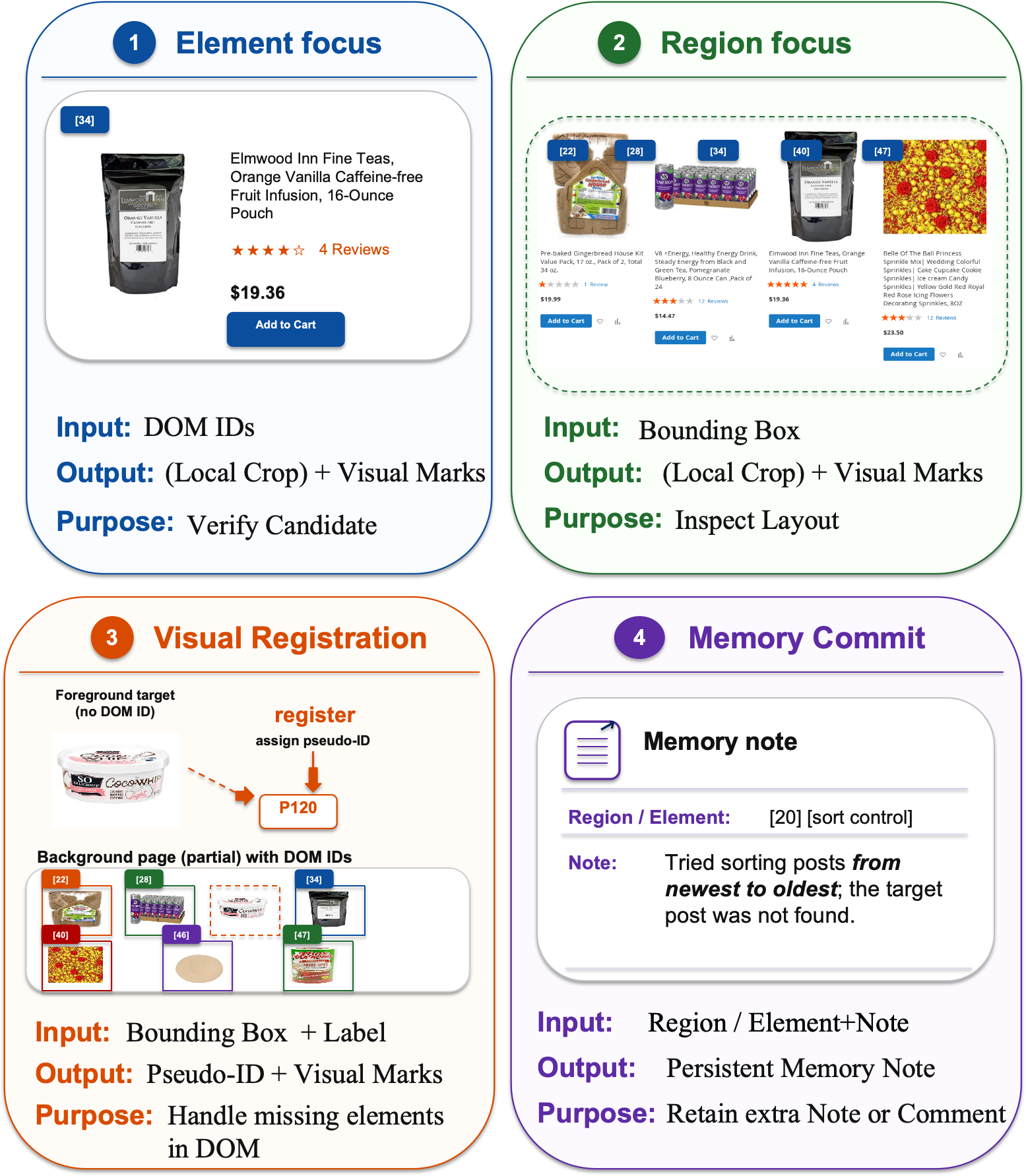}
    \caption{Illustration of the four probing primitives.}
    \label{fig:primitive}
    \vspace{-1.5em}
\end{figure}
The cross-modal bridge is instantiated by three visual grounding probes and one memory probe (Fig.~\ref{fig:overview}). P2A turns cross-modal uncertainty into localized evidence: DOM candidates are rendered onto the screenshot for verification, visual regions are mapped or registered as actionable handles, and probing beliefs are retained across steps. Please refer to Appendix \ref{sec:action_space_prompts} for the prompt schema.
\textbf{(1) Element} and \textbf{(2) Region Focus.}
These two primitives implement the bidirectional DOM--pixel bridge. Element Focus starts from a DOM entry: given an element ID from the text metadata, it renders the element's bounding box on the screenshot and optionally extracts a high-resolution crop, letting the agent check whether the textual description matches what appears on-screen (Fig. \ref{fig:primitive}). Region Focus operates in the reverse direction: given a bounding-box region on the screenshot, it retrieves and highlights the DOM elements falling within that area---useful when the visual layout (e.g., a product grid) is clear but the underlying text structure is not. Together, they realize the on-demand alignment that replaces the static global overlay.
\textbf{(3) Visual Registration}
handles ``ghost elements''---interactable regions (canvas buttons, dynamic overlays) that are missing from the parsed DOM. The agent draws a bounding box around the target and assigns it a pseudo-ID, dynamically extending the symbolic action space so that subsequent operations can reference it by ID.
\textbf{(4) Memory Commit}
binds a short textual note (e.g., ``price verified: \$19.99'') to a specific element or region. Unlike the other primitives, which augment the \textit{current} observation, commit writes to the \textit{persistent} evidence record, ensuring that high-level conclusions survive page transitions.
\subsection{Evidence-based Memory}
\label{sec:memory}
As noted in Sec. \ref{sec:overview}, probe overlays become stale after a browser operation changes the page. Retaining full observations across steps is prohibitively expensive (Tab. 4), yet discarding all past context forces the agent to operate without memory in partially observable tasks.
We therefore retain only the evidence that directly supported past decisions. Since probing reflects what the agent chose to inspect before acting, probed elements form an informative and decision-faithful history. Specifically, an element's metadata is kept in $H_t$ if it was: (1) inspected during verification via Element Focus, Region Focus, or Visual Registration; (2) the target of the final browser operation $a_t$; or (3) explicitly saved via Memory Commit.

All other page content, including screenshots, navigation chrome, footers, and unvisited DOM regions, is discarded. The resulting history stores elements or regions in the observation that the agent explicitly probed, acted on, or committed as evidence, while reaching only about $1.2\times$ the peak retained input context of a no-observation history, compared with $\sim\!3\times$ for retaining full observations. We keep this pruned observation history for the last $N$ steps, with $N{=}5$ by default.

This turns context management from a post-hoc compression problem into a by-product of active perception: the agent decides what to remember by deciding what to inspect.
\subsection{Distilling Active Probing into Open-Weight Models}
\label{sec:distillation}
Proprietary models such as Gemini-3-Pro can leverage P2A zero-shot via prompting. Open-weight models (e.g., Qwen3-VL-8B), however, lack the capacity to spontaneously plan probe sequences. We bridge this gap with a two-stage distillation pipeline as follows. More details and statistics are provided in Appendix \ref{sec:data_pipeline}

\paragraph{Cold-Start Data Synthesis.}
We construct an initial training corpus through a four-step pipeline:
\textbf{(a) Task Synthesis.} We use Gemini-3-Flash to generate diverse browser tasks on random pages from the VWA environment via self-instruction~\cite{pahuja2025explorer,he2025openwebvoyager} without any prior knowledge besides the website environments, and enable vision-dependent tasks by allowing the generator to reference image URLs in the DOM.
\textbf{(b) Teacher Rollout.} A strong teacher (Gemini-3-Flash) performs single-step P2A rollouts on truncated trajectories from step (a). We retain only those rollouts whose resulting browser operation matches the original successful action.
\textbf{(c) Rejection Sampling.} A judge model verifies task success from the final page state~\cite{deng2023mind2web,qi2024webrl}; unsuccessful trajectories are discarded.
\textbf{(d) Data Mixing and Balancing.} We mix filtered probing trajectories with GUI grounding samples~\cite{liu2025scalecua} and pure browser-operation trajectories, and add extra targeted rollouts~\cite{he2025openwebvoyager,qi2024webrl} to cover long-tail \texttt{Memory Commit} and \texttt{Visual Registration} cases.

\paragraph{Multi-Round Self-Bootstrapped SFT.}
Starting from the cold-start checkpoint, we iteratively roll out the current model, filter successful trajectories with a model-based correctness judge, mix them with existing training datasets for training, and fine-tune the next checkpoint. Performance improves across rounds while probe usage becomes more selective (Sec.~\ref{sec:ablation}).

\newcommand{\mainvwatable}{%
\begin{table*}[!t]
\centering
\caption{Performance Comparisons with both DOM-based and Screenshot-based browser-use agents on VisualWebArena across different sites.}
\label{tab:vwa_results_split}
\begin{adjustbox}{max width=0.95\textwidth}
\begin{tabular}{@{}lcccccccc@{}}
\toprule
\multirow{2}{*}{\textbf{Method}} &
\multirow{2}{*}{\textbf{Obs.}} &
\multirow{2}{*}{\textbf{Zero Shot}} &
\multicolumn{4}{c}{\textbf{Success Rate $\uparrow$}} \\
\cmidrule(lr){4-6}
& & & \textbf{Classifieds} & \textbf{Shopping} & \textbf{Reddit} & \textbf{Avg.} \\
\midrule

ScaleCUA-7B~\cite{liu2025scalecua} & {\small Screenshot} & \cmark & 11.1 & 16.7 & 16.6 & 15.3\\
ScaleCUA-32B~\cite{liu2025scalecua}           & {\small Screenshot} & \cmark & 26.1 & 24.7 & 20.5 & 24.1\\
Qwen3-VL-8B~\cite{bai2025qwen3}             & {\small Screenshot} & \cmark & 15.8 & 19.1 & 20.0 & 18.5\\

\midrule
GPT-4o~\cite{koh2024visualwebarena}      & {\small DOM+SoM} & \cmark & 17.9 & 14.3 & 19.3 & 17.6\\
GPT-5.2~\cite{openai2025gpt52}               & {\small DOM+SoM} & \cmark & 48.3 & 49.1 & 37.6 & 46.3\\
SGV~\cite{andrade2025let}                & {\small DOM+SoM} & \cmark & 52.0 & 57.0 & 33.0 & 50.2\\
WALT \cite{prabhu2025walt}                                     & {\small DOM+SoM} & \cmark & 64.1 & 53.4 & 39.0 & 52.9\\

\midrule
Gemini-2.5-Flash~\cite{GoogleDeepMind2025Gemini25}   & {\small DOM+SoM} & \cmark & 38.6 & 42.0 & 32.9 & 38.9\\
\rowcolor{blue!5}
+ P2A (Prompt)                            & {\small DOM+SoM} & \cmark & 39.0 & 44.0 & 35.9 & 40.4\\

\midrule
Gemini-3-Flash~\cite{googleflash}     & {\small DOM+SoM} & \cmark & 49.1 & 41.4 & 37.1 & 42.4\\
\rowcolor{blue!5}
+ P2A (Prompt)                            & {\small DOM+SoM} & \cmark & 50.2 & 50.8 & 39.0 & 48.7\\

\midrule
Gemini-3-Pro~\cite{googlepro}         & {\small DOM+SoM} & \cmark & 64.6 & 51.2 & \textbf{48.7} & 54.1\\
\rowcolor{blue!5}
+ P2A (Prompt)                            & {\small DOM+SoM} & \cmark & \textbf{80.0} & \textbf{59.0} & 45.0 & \textbf{61.2}\\
\midrule
\midrule
Qwen3-VL-8B \cite{bai2025qwen3}                               & {\small DOM+SoM} & \cmark & 27.8 & 29.6 & 10.0 & 24.6\\
+ Operation Cold-Start                    & {\small DOM+SoM} & \xmark & 36.3 & 16.5 & 19.5 & 22.3\\
+ P2A (Prompt)                             & {\small DOM+SoM} & \cmark & 23.1 & 19.3 & 9.5  & 18.0\\
\rowcolor{blue!5}
+ P2A (Cold-Start)                         & {\small DOM+SoM} & \xmark & 37.2 & 31.6 & 17.4 & 29.9\\
\rowcolor{blue!5}
+ P2A (Round 1)                       & {\small DOM+SoM} & \xmark & 42.3 & 30.7 & 19.0 & 31.0\\
\rowcolor{blue!5}
+ P2A (Round 2)                       & {\small DOM+SoM} & \xmark & \textbf{44.4} & \textbf{32.8} & \textbf{20.0} & \textbf{32.9}\\

\midrule
\textcolor{gray}{Human}~\cite{koh2024visualwebarena} &
\textcolor{gray}{{\small --}} &
\textcolor{gray}{{\small --}} &
\textcolor{gray}{91.7} & \textcolor{gray}{88.4} & \textcolor{gray}{87.1} & \textcolor{gray}{\textbf{88.7}}\\

\bottomrule
\end{tabular}
\end{adjustbox}
\vspace{-1em}
\end{table*}%
}
\section{Experiments}
\subsection{Setup}
\paragraph{Benchmark and Environment.}
We evaluate \textit{Probe to Act} primarily on \textbf{VisualWebArena (VWA)}~\cite{koh2024visualwebarena}, a standard benchmark of 910 multimodal browser-use tasks across three domains (Classifieds, Reddit, Shopping) that stresses image-conditioned reasoning and long-horizon navigation under rigorous functional oracles rather than LLM judges. We further report \textbf{WebArena (WA)}~\cite{zhou2023webarena}, which shares VWA's web ecosystem, and \textbf{Online-Mind2Web}~\cite{xueillusion}, which tests navigation on open, live websites. Following~\cite{andrade2025let}, we adopt the corrected evaluator protocols on VWA and WA and the curated \texttt{VWA-Lite} subset (305 tasks) for ablations. o4-mini \cite{OpenAI2025o3o4mini} serves as the judge model of Online-Mind2Web tasks. For reproducibility, we locally deploy the 7 offline websites shared by the VWA and WebArena ecosystems in a self-contained Docker, used for both rollout and evaluation.
\mainvwatable
\newcommand{\supporttables}{%
\begin{table*}[!t]
\centering
\vspace{-0.5em}

\begin{minipage}[t]{0.38\textwidth}
\centering

\caption{\small Performance of P2A on WebArena and Online-Mind2Web.}
\label{tab:cross_bench}
\setlength{\tabcolsep}{3.5pt}
\renewcommand{\arraystretch}{1.03}
\scriptsize
\resizebox{\linewidth}{!}{%
\begin{tabular}{@{}lcc@{}}
\toprule
\textbf{Method} & \textbf{WebArena} & \textbf{Online-Mind2Web} \\
\midrule
Gemini-2.5-Flash & 34.5 & 46.7 \\
\rowcolor{blue!5}
\quad + P2A & 37.2 & 48.3 \\
\midrule
Gemini-3-Flash & 47.7 & 49.0 \\
\rowcolor{blue!5}
\quad + P2A & \textbf{51.0} & \textbf{51.7} \\
\midrule
Qwen3-VL-8B & 18.2 & 21.3 \\
\rowcolor{blue!5}
\quad + P2A & 24.3 & 25.7 \\
\bottomrule
\end{tabular}%
}

\vspace{0.45em}

\caption{\small Ablation study of Probing Primitives on VWA-Lite.}
\label{tab:ablation_primitives}
\setlength{\tabcolsep}{4pt}
\renewcommand{\arraystretch}{1.03}
\scriptsize
\resizebox{\linewidth}{!}{%
\begin{tabular}{cccc|c}
\toprule
\textbf{Element} & \textbf{Region} & \textbf{Visual} & \textbf{Memory} & \textbf{Success} \\
\textbf{Focus} & \textbf{Focus} & \textbf{Register} & \textbf{Commit} & \textbf{Rate} \\
\midrule
\checkmark &            &            &            & 29.8 \\
          & \checkmark  &            &            & 27.3 \\
\checkmark & \checkmark &            &            & 31.6 \\
\checkmark & \checkmark & \checkmark &            & 31.9 \\
\rowcolor{blue!5}
\checkmark & \checkmark & \checkmark & \checkmark & \textbf{32.6} \\
\bottomrule
\end{tabular}%
}

\end{minipage}
\hfill
\begin{minipage}[t]{0.58\textwidth}
\centering

\caption{\small Ablation of evidence-based memory on VWA-Lite. ``Peak Ctx.'' denotes the maximum retained input-token count relative to the Action-Only baseline. ``Action-centric'' retains each operated element together with the $K$ preceding and following DOM rows.}
\label{tab:ablation_memory}
\setlength{\tabcolsep}{4.2pt}
\renewcommand{\arraystretch}{1.03}
\scriptsize
\resizebox{\linewidth}{!}{%
\begin{tabular}{@{}llcc@{}}
\toprule
\textbf{Model} & \textbf{Retention Strategy} 
& \textbf{Peak Ctx.} 
& \textbf{Success} \\
& & \textbf{($\times$)} & \textbf{(\%)} \\
\midrule
\multirow{7}{*}{\textbf{Qwen3-VL-8B}}
& Action-Only & 1.0 & 24.6 \\
& Full History & 2.8 & 27.4 \\
& \cellcolor{blue!5}\textbf{Ours w/o obs history}
& \cellcolor{blue!5}\textbf{1.1}
& \cellcolor{blue!5}30.3 \\
& \cellcolor{blue!5}\textbf{Ours action-centric ($K{=}2$)}
& \cellcolor{blue!5}1.2
& \cellcolor{blue!5}31.0 \\
& \cellcolor{blue!5}\textbf{Ours action-centric ($K{=}4$)}
& \cellcolor{blue!5}1.35
& \cellcolor{blue!5}30.6 \\
& \cellcolor{blue!5}\textbf{Ours + full observation}
& \cellcolor{blue!5}3.2
& \cellcolor{blue!5}\textbf{33.7} \\
& \cellcolor{blue!5}\textbf{Ours evidence memory}
& \cellcolor{blue!5}1.2
& \cellcolor{blue!5}32.6 \\
\midrule
\multirow{5}{*}{\textbf{Gemini-3-Flash}}
& Action-Only & 1.0 & 42.7 \\
& Full History & 2.9 & 47.9 \\
& \cellcolor{blue!5}\textbf{Ours w/o obs history}
& \cellcolor{blue!5}\textbf{1.1}
& \cellcolor{blue!5}46.1 \\
& \cellcolor{blue!5}\textbf{Ours + full observation}
& \cellcolor{blue!5}3.2
& \cellcolor{blue!5}\textbf{49.1} \\
& \cellcolor{blue!5}\textbf{Ours}
& \cellcolor{blue!5}\textbf{1.3}
& \cellcolor{blue!5}48.7 \\
\bottomrule
\end{tabular}%
}

\end{minipage}

\vspace{-0.8em}
\end{table*}%
}
\paragraph{Baselines.}
We compare against baselines under three configurations: \textbf{(1) Standard DOM-SoM}, following the community-standard VWA protocol adopted by the majority of prior work, where agents receive a parsed text DOM and a SoM-overlaid screenshot. We evaluate the Gemini series \cite{googlepro,googleflash,GoogleDeepMind2025Gemini25}, GPT-5.2\cite{openai2025gpt52}, and Qwen3-VL \cite{bai2025qwen3} under this shared protocol; closed-source proprietary computer-use products such as OpenAI Operator\cite{openai2025operator} whose internals are undisclosed do not constitute reproducible baselines under a controlled interface and are therefore excluded. \textbf{(2) Screenshot-Coordinate}, where agents operate on raw screenshots via pixel coordinates; we evaluate ScaleCUA \cite{liu2025scalecua} and Qwen3-VL with zero-shot grounding. \textbf{(3) Operation-Only Control}: Qwen3-VL-8B fine-tuned on our cold-start data with all probing steps removed, denoted \textit{Operation Cold-Start}. All evaluations are zero-shot with a budget of 30 steps unless noted. On WebArena and Online-Mind2Web we evaluate representative baselines of proprietary and distilled settings.
\paragraph{Implementation Details.}
We evaluate P2A in a training-free setting using the Gemini series and fine-tune Qwen3-VL-8B \cite{bai2025qwen3} to demonstrate learnability on an open-weight model.
For cold-start synthesis, we employ Gemini-3-Flash \cite{googleflash} to generate tasks via self-instruction and collect pure operation trajectories. We curate reasoning data from 1,000 tasks, retaining Probe-to-Act trajectories only when they yield operations equivalent to successful pure rollouts.
This dataset is augmented with 80k GUI grounding samples from ScaleCUA \cite{liu2025scalecua} and an additional 1000 pure-operation task trajectories.
For Self-Bootstrapped SFT, we rely solely on the currently trained model to collect trajectories.
Throughout the pipeline, Gemini-3-pro \cite{googlepro} serves as the outcome verifier.
Unless otherwise specified, we cap the reasoning budget at 10 probing steps per browser operation. For WebArena and Online-Mind2Web, we apply the same prompt and inference configuration 
as VWA with no benchmark-specific adaptation.
Please refer to the Appendix \ref{sec:implementation} for further details.
\subsection{Main Results}
Tables~\ref{tab:vwa_results_split} and~\ref{tab:cross_bench} show the main results.  %
In the training-free setting, equipping proprietary models with probing yields consistent gains: Gemini-3-Flash rises from 42.4\% to 48.7\% on VWA, and the same prompt harness carries over to WebArena and Online-Mind2Web with at most marginal regression---evidence that strong reasoners immediately benefit from active visual probing. Conversely, Qwen3-VL-8B fails to leverage P2A zero-shot, exhibiting limited ability to align dynamic visual prompts with textual layouts without supervision and motivating our distillation pipeline.

Our distillation lifts Qwen3-VL-8B to 32.9\% after two self-bootstrapping rounds on VWA, with the same upward trend echoed on WebArena and Online-Mind2Web. At the matched Cold-Start stage, operation-only supervision reaches 22.3\%, whereas interface-aligned probe--evidence supervision reaches 29.9\%; the subsequent gain to 32.9\% is attributed to the complete P2A self-bootstrapping pipeline. Appendix Figure~\ref{fig:success_vs_steps} further shows a higher success rate with fewer operations, while Appendix \ref{sec:qualitative examples} provides qualitative examples.

\subsection{Ablation Study}
\label{sec:ablation}
\supporttables
\begin{figure*}[!t]
    \centering
    \begin{minipage}[t]{0.35\textwidth}
        \centering
        \includegraphics[width=\linewidth]{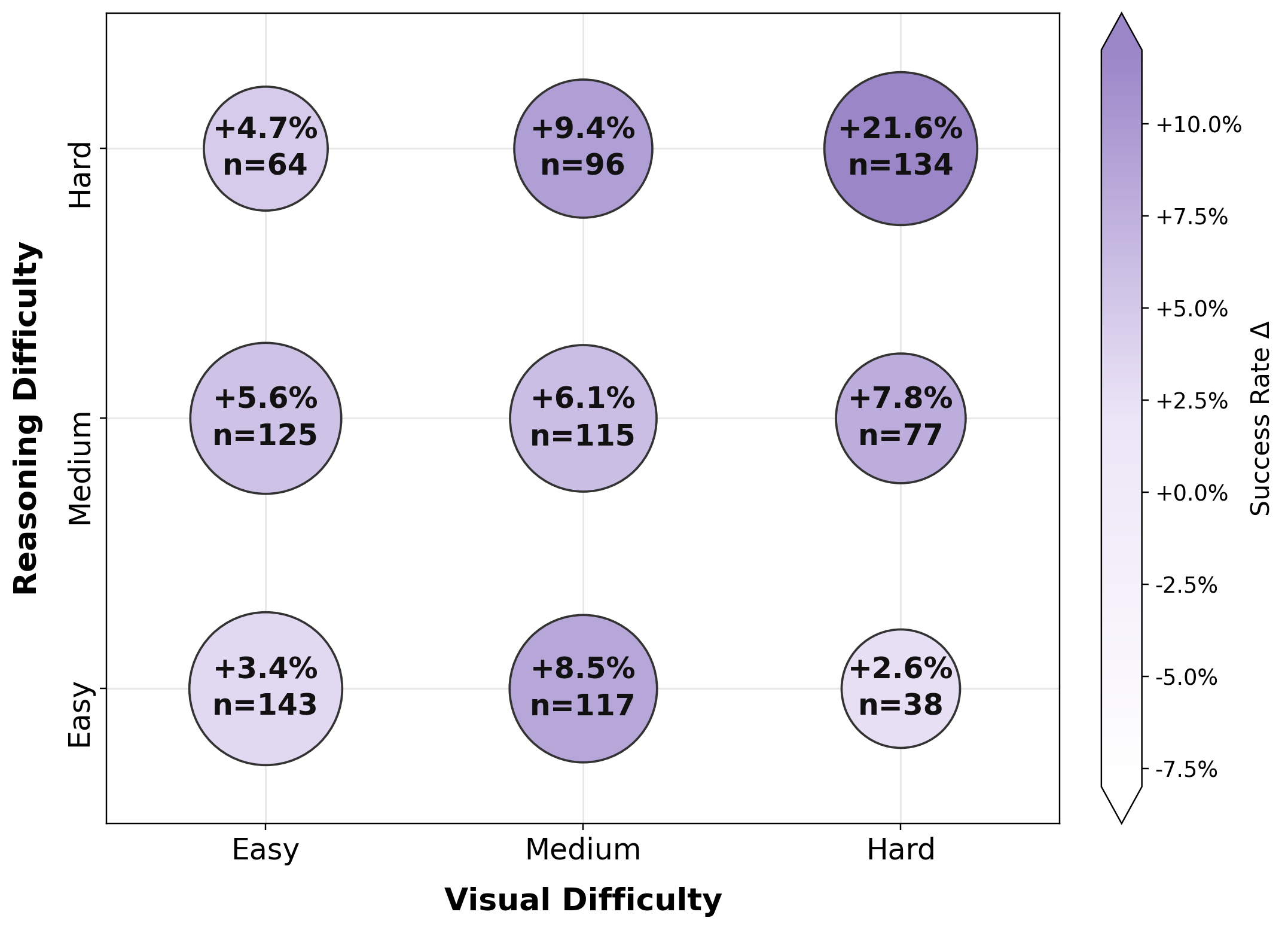}
        \caption{Grouped performance gain of the P2A Round 2 tuned model over the baseline Qwen3-VL-8B.
        }
        \label{fig:difficulty_delta}
    \end{minipage}
    \hspace{0.02\textwidth}
    \begin{minipage}[t]{0.55\textwidth}
        \centering
        \includegraphics[width=\linewidth]{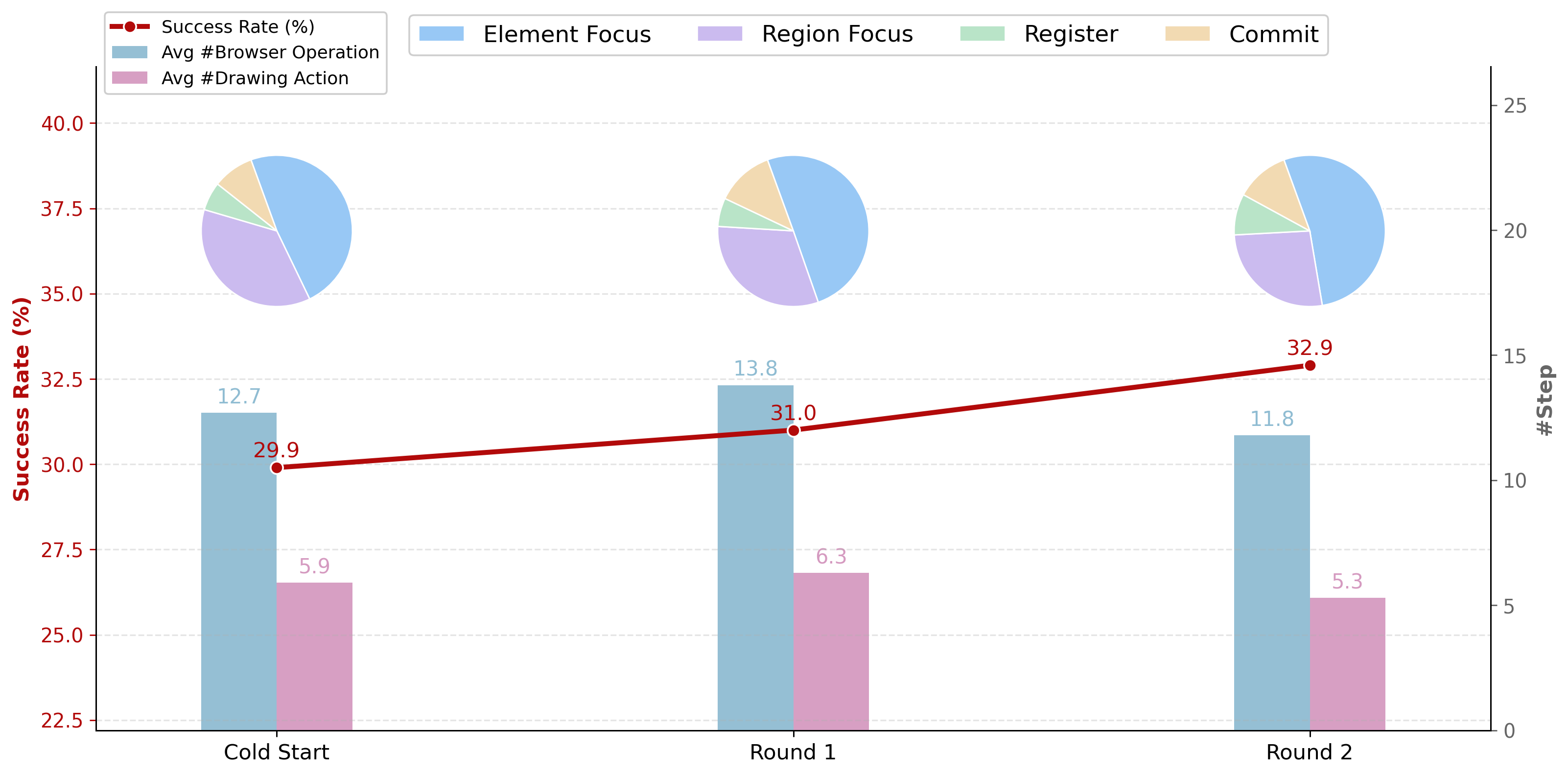}
        \caption{Performance dynamics, including success rate, average number of operations and probing actions, and frequencies across probing primitives, are shown across SFT rounds.}
        \label{fig:training_dynamics}
    \end{minipage}
    \vspace{-0.8em}
\end{figure*}
\paragraph{Probed Observation as Efficient Memory.}
We compare \textit{Evidence-based Memory} with Action-Only history, which discards past observations, and Full-History retention, which keeps the complete DOM within a sliding window (Table~\ref{tab:ablation_memory}). Active probing itself provides the main gain: even without cross-step observation history, P2A reaches 30.3\%, compared with 24.6\% for the standard baseline. Evidence memory then improves the accuracy--context trade-off by retaining only the elements that informed the agent's decisions. It reaches 32.6\% at $1.2\times$ peak retained input context, outperforming standard Full History at substantially lower context cost and approaching the accuracy of retaining full observations with P2A.

To test whether this benefit can be recovered from executed actions alone, we additionally consider an \emph{action-centric} heuristic that stores each operated element together with the $K$ preceding and following DOM rows. With $K{=}2$, this heuristic uses the same $1.2\times$ context budget but reaches 31.0\%, below evidence memory; expanding the neighborhood to $K{=}4$ increases context without improving accuracy. Thus, the evidence selected by pre-action probes and explicit commits contains decision-relevant information that is not reliably recoverable from a fixed neighborhood around operated elements. This supports our view of memory as a co-product of reasoning rather than a separate post-hoc compressor.
\paragraph{Ablation on Primitives.}
We analyze the contribution of individual probing primitives on the \texttt{VWA-Lite} subset in Table~\ref{tab:ablation_primitives}.
\textbf{Element Focus} alone yields a significant performance gain, outperforming \textbf{Region Focus} (+5.2\% vs. +2.7\%). This highlights the necessity of textual metadata for efficient page understanding and candidate selection, whereas Region Focus relies heavily on the model's intrinsic raw localization capabilities.
Crucially, combining both verification mechanisms further boosts the success rate to 31.6\%, highlighting a synergy between semantic and spatial perspectives.

\subsection{Further Analysis}
\paragraph{Inference Cost.}
The 30-step budget counts state-changing browser operations only, while all probe turns are included in the inference accounting. Table~\ref{tab:inference_accounting_main} contrasts trained and prompt-based P2A. After training, P2A improves Qwen success while reducing browser operations by nearly one third, with only a moderate cost increase. In contrast, prompt-only P2A on Qwen consumes substantially more inference time but performs worse, showing that additional computation alone is insufficient. Prompt-based P2A on Gemini improves success with a different cost profile dominated by additional image inputs. We therefore characterize P2A as additional within-step deliberation rather than a fixed-compute improvement; Appendix Table~\ref{tab:inference_accounting} further reports the corresponding probe-turn counts.
\begin{table}[t]
\centering
\caption{Representative average per-task inference accounting on VWA. ``SR'' denotes task success rate, and ``Ops'' counts state-changing browser operations under the shared 30-step budget. Image inputs and wall time include all within-step probe turns.}
\label{tab:inference_accounting_main}
\scriptsize
\setlength{\tabcolsep}{3.2pt}
\renewcommand{\arraystretch}{1.05}
\resizebox{\columnwidth}{!}{%
\begin{tabular}{@{}lrrrr@{}}
\toprule
\textbf{Model} & \textbf{SR (\%)} & \textbf{Ops} & \textbf{Images} & \textbf{Time} \\
\midrule
Qwen3-VL-8B-Instruct & 24.6 & 15.84 & 16.04 & 203.2 s \\
\quad + P2A (Round 2) & \textbf{32.9} & 11.25 & 16.90 & 243.8 s \\
\quad + P2A (Prompt) & 18.0 & 19.24 & 20.83 & 338.6 s \\
\midrule
Gemini-3-Flash & 42.4 & 12.80 & 13.20 & 380.1 s \\
\quad + P2A (Prompt) & \textbf{48.7} & 13.08 & 20.70 & 432.8 s \\
\bottomrule
\end{tabular}%
}
\end{table}
\paragraph{Robustness to Missing Evidence.}
We further isolate Visual Registration and Memory Commit through controlled randomized evidence-loss tests. At 30\% DOM removal, full P2A achieves an 11.1\% SR gain over the standard DOM+SoM baseline and a 4.8\% SR gain over the no-Registration variant, supporting Registration's role in recovering visually present but structurally unavailable targets. At 50\% history removal, full P2A achieves an 8.3\% SR gain over the baseline and a 2.0\% SR gain over the no-Commit variant, supporting Commit's role in preserving compact conclusions across steps. The Commit gain remains positive under heavier deletion, although information removed before it can be summarized or committed is no longer recoverable. Appendix Table~\ref{tab:missing_evidence_robustness} further reports the clean-condition results and changes relative to clean for all settings.
\begin{table}[t]
\centering
\caption{Controlled robustness to randomized missing evidence on VWA. Interactable DOM records and their SoM marks are randomly removed while pixels and true interactivity are preserved; complete prior history turns are randomly removed while the current screenshot, DOM, and browser state are preserved. ``Baseline'' denotes the standard DOM+SoM agent. All entries are task success rates (\%) except ``Gain,'' which compares full P2A with the corresponding primitive ablation.}
\label{tab:missing_evidence_main}
\scriptsize
\setlength{\tabcolsep}{10pt}
\renewcommand{\arraystretch}{1.05}
\begin{adjustbox}{max width=\columnwidth}
\begin{tabular}{@{}rrrrr@{}}
\toprule
\multicolumn{5}{@{}l}{\textcolor{gray}{\scriptsize Random DOM Element Removed}} \\
\midrule
\shortstack{\textbf{Removed}\\\textbf{Ratio}} &
\textbf{Baseline} &
\textbf{P2A} &
\shortstack{\textbf{W/o}\\\textbf{Registration}} &
\textbf{Gain} \\
\midrule
10\% & 22.4 & 31.0 & 26.4 & +4.6 \\
30\% & 16.0 & 27.1 & 22.3 & +4.8 \\
\midrule
\multicolumn{5}{@{}l}{\textcolor{gray}{\scriptsize Random History Turn Removed}} \\
\midrule
\shortstack{\textbf{Removed}\\\textbf{Ratio}} &
\textbf{Baseline} &
\textbf{P2A} &
\shortstack{\textbf{W/o}\\\textbf{Commit}} &
\textbf{Gain} \\
\midrule
30\% & 19.2 & 28.0 & 25.7 & +2.3 \\
50\% & 18.0 & 26.3 & 24.3 & +2.0 \\
\bottomrule
\end{tabular}
\end{adjustbox}
\end{table}
\paragraph{Improvement on Difficulty Splits.}
Figure \ref{fig:difficulty_delta} exhibits performance gains over Qwen3-VL-8B across VWA's difficulty splits, categorized by visual and reasoning complexity.
P2A delivers broadly consistent gains across all 9 visual$\times$reasoning strata, with the largest uplift under compounded \textbf{hard-visual$\times$hard-reasoning tasks}. This confirms the benefit concentrates precisely where DOM--pixel alignment is complex, rather than merely easing perception in simple layouts.
\paragraph{Dynamics of Multiple-Turn Tuning.}
We track how the agent's behavior evolves across the three training stages---Cold-Start, Round 1, and Round 2 SFT (Fig.~\ref{fig:training_dynamics}). As success rises monotonically, probing frequency follows an inverted-U, peaking at Round 1 but decreasing after Round 2, while its composition tilts from exploratory \texttt{Region Focus} toward confirmatory \texttt{Element Focus} and \texttt{Register}. Early on, the agent leans on \texttt{Region Focus} more to locate relevant elements from spatial priors; it later pivots to confirming candidates against DOM priors, exhibiting two-way DOM--screenshot bridge via iterative probing.

\section{Conclusion}
We presented P2A, a browser-agent interface that treats DOM--pixel alignment as a decision-time verification process rather than a one-shot static overlay. By interleaving state-preserving probes with reasoning, P2A lets the agent verify candidate DOM elements, inspect visually salient regions, register off-DOM targets, and retain only evidence that was actually inspected or interacted with. Experiments on VisualWebArena, WebArena, and Online-Mind2Web show that this design improves both prompted proprietary models under the DOM+SoM interface and distilled open-weight models, while evidence-based memory achieves a favorable accuracy--context tradeoff compared with full-observation history. Our qualitative analysis suggests that the gains come from both more deliberate action selection and a memory trace tied to the agent's own verification process. Future work would explore adaptive probing budgets, enrich probes with more visual tools, and extend the framework to general UI modalities.

\section*{Limitations}
While the proposed method shows consistent improvements across multiple browser-use benchmarks, our evaluation remains centered on web navigation scenarios within the DOM+SoM interface. Extending the framework to broader interactive environments, such as desktop or mobile UI agents, remains an important direction for future work. In addition, the current implementation uses a fixed set of lightweight probing primitives and a bounded probing budget. More flexible probe selection and richer visual tools may further improve efficiency and generality. Finally, although our evidence-based memory provides a favorable accuracy--context trade-off, future work may explore more flexible memory policies for interaction trajectories in more diverse and complex scenes. Our Qwen3-VL-8B study demonstrates learnability on one open-weight model family rather than a scaling trend.
\bibliography{main}

\appendix

\clearpage
\begin{table*}[ht]
\centering
\caption{Summary of notation used in the Probe-to-Act formulation.}
\label{tab:notation_summary}
\footnotesize
\setlength{\tabcolsep}{4pt}
\renewcommand{\arraystretch}{1.05}
\begin{tabularx}{\textwidth}{@{}>{\raggedright\arraybackslash}p{0.23\textwidth}X@{}}
\toprule
\textbf{Symbol} & \textbf{Meaning} \\
\midrule
\(t\) & Outer browser-use decision step; each step ends with one browser operation. \\
\(d_t, v_t, o_t\) & DOM text, screenshot, and raw multimodal observation \(o_t=(d_t,v_t)\) at step \(t\). \\
\(s, q, H_{t-1}\) & System prompt, user instruction, and inter-step history available before step \(t\). \\
\(\pi_\theta\) & Agent policy parameterized by model parameters \(\theta\). \\
\(\mathcal{A}_{\text{probe}}\) & State-preserving probing action space used to augment the agent's observation. \\
\(\mathcal{A}_{\text{browser}}\) & State-changing browser-operation action space, e.g., click, type, scroll, or navigate. \\
\(k, K, p_{t,k}\) & Probe index, total number of probes in step \(t\), and the \(k\)-th probe action \(p_{t,k}\in\mathcal{A}_{\text{probe}}\). \\
\(o_{t,k}\) & Internal observation after \(k\) probe updates, with \(o_{t,0}=o_t\) and \(o_{t,k}=\mathrm{apply}(p_{t,k}, o_{t,k-1})\). \\
\(r_{t,k}\) & Textual reasoning segment associated with the \(k\)-th internal observation; \(r_{t,K}\) precedes the final operation. \\
\(a_t\)  & Final browser operation at step \(t\) \\
\(\tau_t\) & Step-local probe-act trajectory \((r_{t,0}, p_{t,1}, r_{t,1}, \ldots, p_{t,K}, r_{t,K}, a_t)\). \\
\(M_t\) & Sparse evidence memory retained after step \(t\) from probed, acted-on, or committed evidence. \\
\bottomrule
\end{tabularx}
\end{table*}
\section{Notation Summary}
\label{sec:notation_summary}
For clarity, we summarize the main notations used in Sec.~\ref{sec:method} in
Table~\ref{tab:notation_summary}.

\begin{table*}[ht]
\centering
\caption{Statistics of the data generation and multi-round rejection sampling pipeline. \textit{Gen.} indicates the generated data before filtering, while \textit{Retained} denotes the high-quality data not rejected, kept for SFT. \textit{Probe \%} represents the proportion of internal probing primitive turns among all action turns.}
\label{tab:data_statistics}
\resizebox{\textwidth}{!}{
\begin{tabular}{llcccc}
\toprule
\textbf{Stage} & \textbf{Data Type / Source} & \textbf{\# Gen. Traj.} & \textbf{\# Gen. Turns (Probe \%)} & \textbf{\# Retained Turns (Probe \%)} & \textbf{Rejection Rate (\%)} \\
\midrule
\multirow{3}{*}{\textbf{Cold Start}} 
& Operation w/ Probing Primitives          & 3375 & 41587(36.6 \%) &  24052 (42.9 \%) & 42.2 \% \\
& Pure Operation    & 1943 & 21995 & 11771 & 46.5 \% \\
& GUI Grounding           & $-$ & $-$ & 40000 & $-$  \\
\midrule
\multirow{1}{*}{\textbf{Round 1 Self-Bootstrapped}} 
& Operation w/ Probing Primitives     & 2260 & 42985 (37.3 \%) & 17931 (39.4 \%) & 58.3 \% \\
\midrule
\multirow{1}{*}{\textbf{Round 2 Self-Bootstrapped}} 
&  Operation w/ Probing Primitives     & 1497 & 27245 (36.8 \%) & 13606 (41.3 \%) & 50.0 \% \\
\bottomrule
\end{tabular}
}
\end{table*}

\section{Data Generation Pipeline \& Statistics}
\label{sec:data_pipeline}
\subsection{Data Statistics \& Distribution}
Table \ref{tab:data_statistics} summarizes the data composition and filtering statistics across our multi-stage training pipeline: Cold-Start synthesis, Round 1 Self-Bootstrapping, and Round 2 Self-Bootstrapping. To guarantee the quality of the probing-augmented reasoning traces, we apply strict trajectory-level rejection sampling via a Judge VLM. We report the number of generated trajectories, the total action turns before and after filtering, and the corresponding rejection rates. During self-bootstrapping, the newly retained data is mixed with downsampled data from the previous round at a 70:30 ratio. Furthermore, the proportion of probing primitives (\textit{Probe \%}) within the action turns is provided to illustrate the density of internal visual reasoning at each stage.

\begin{figure}[!htbp]
    \centering
    \includegraphics[width=\linewidth]{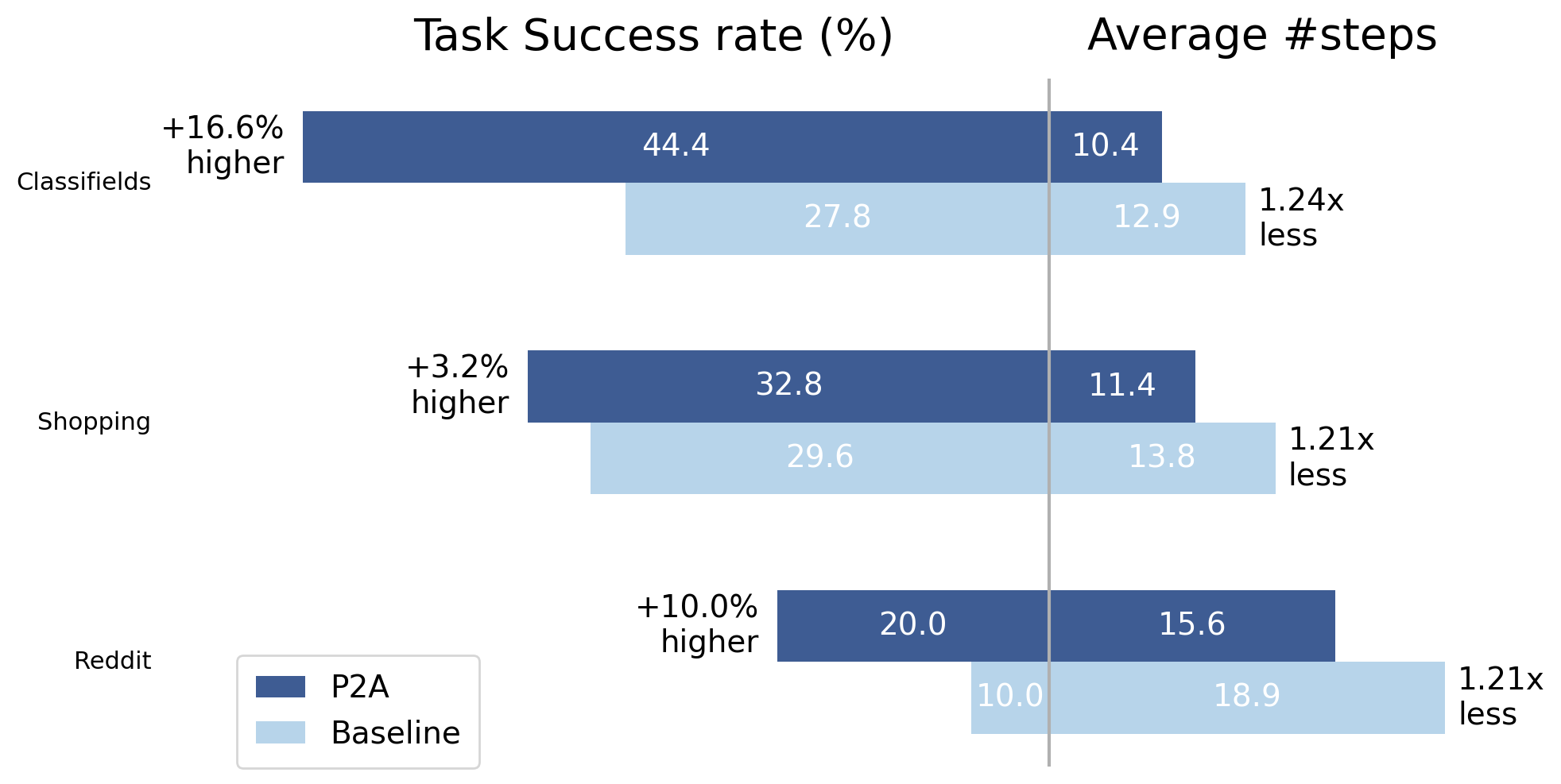}
    \caption{We compare our framework's performance and operation steps across three sites of VWA, with the baseline Qwen3-VL-8B as control.}
    \label{fig:success_vs_steps}
\end{figure}

\paragraph{Per-Site Success--Efficiency Trade-off.}
Appendix Figure~\ref{fig:success_vs_steps} compares P2A with the Qwen3-VL-8B baseline on Classifieds, Reddit, and Shopping. Across all three VWA sites, P2A shifts the points toward higher success with fewer browser-operation steps, showing that the gain is not driven by longer interaction traces. This pattern aligns with the main-text observation that probing encourages the agent to verify candidates before committing to browser actions. By catching ambiguous or visually unsupported choices earlier, the agent avoids premature wrong operations and therefore needs fewer recovery steps.

\subsection{Synthesis Task Proposal}
\label{sec:task_proposal}

To synthesize task queries for browser-use agents for the training corpus, we employ the self-construct approach from Explorer \cite{pahuja2025explorer}, where an agent freely explores the web environment using the parsed textual DOM and Set-of-Marks (SoM) screenshots. The agent iteratively refines its navigation goal online based on its actions, and a feasible task is subsequently summarized from the finalized trajectory. To explicitly generate multimodal, image-conditioned browser tasks, we embed the source URLs of webpage image objects directly into the agent's visible DOM. During exploration, the agent is specifically encouraged to use these visual objects as task conditions and reference them via their URLs when updating its goal. Crucially, during the subsequent trajectory collection phase, these referenced URLs are processed into actual image inputs, thereby naturally yielding true multimodal tasks. The exact prompts used for the navigation and task refinement stage, as well as the final task summary stage, are detailed in Figure~\ref{fig:task_proposal_prompts}.
\vspace{1em} %
\noindent
\begin{tcolorbox}[
    enhanced,            %
    breakable,           %
    colback=gray!5!white, 
    colframe=gray!50!black, 
    title=\textbf{Prompts for Synthesis Task Proposal}, 
    fonttitle=\small\bfseries, 
    arc=2mm, 
    boxrule=0.5pt,
    width=\linewidth,    %
    left=4mm,            %
    right=4mm,           %
    top=3mm,             %
    bottom=3mm           %
]

\small
\textbf{Stage 1: Navigation and Task Refinement Prompt} \\
\rule{\linewidth}{0.4pt}
\begin{Verbatim}[breaklines=true]
Imagine you are a real user on this webpage, and your overall task is {overall_task}. This is the list of actions you have performed that led to the current page {prev_action_list}. You are also given the webpage screenshot and parsed HTML/accessibility tree.
Do the following step by step:
1. Please predict what action the user might perform next that is consistent with the overall task and previous action list in natural language.
2. Then based on the parsed HTML/accessibility tree of the webpage and the natural language action, generate the grounded action.
3. Update the overall task aligned with this set of actions.
4. When appropriate, refine the task into an image-conditioned one: if the parsed HTML/accessibility tree exposes image URLs for images that are visibly present in the screenshot (e.g., product photos, post thumbnails, profile avatars), you are encouraged to use that image URL or a distinctive substring/path from it as a condition for identifying the target.

*  Task update rules *
1. The task must contain some actions: Buy, Book, Find, Check, Choose show me, give me, add to cart, ...
2. You should only propose tasks that do not require login to execute the task.
3. You should propose tasks that are clear and specific.
4. Update the details of the task, such as price, date, location, etc. based on the current set of actions and the proposed action.
5. When possible, diversify the refined task by using a visible image as the key condition. If the DOM exposes the corresponding image URL, you may identify the target by that image URL or by a distinctive substring/path from it.
6. Any image URL you mention must correspond to an image that is actually visible in the screenshot/current page or clearly grounded by the current valid context. Do NOT invent or hallucinate image URLs.
7. If the updated task uses an image URL condition, the next natural language action should explicitly mention using that image URL cue to locate or interact with the target.

*ACTION SPACE*: {ACTION_SPACE}.

*  Action generation rules *
1. You should generate a single atomic action at each step.
2. The action should be an atomic action from the given action space.
3. The arguments to each action should be within square braces.
4. The natural language form of action should be consistent with the grounded version of the action.
5. If the type action is selected, the natural language form of action should always specify the actual text to be typed.
6. You should issue a stop action if the current webpage asks to login or for credit card information.
7. To input text, there is NO need to click textbox first, directly type content.
8. STRICTLY Avoid repeating the same action if the webpage remains unchanged.
9. Do NOT use quotation marks in the action generation.
10. If the task uses an image URL condition to identify the target, the natural language action should explicitly mention that image URL cue.
\end{Verbatim}

\vspace{2mm}
\textbf{Stage 2: Task Summary Prompt} \\
\rule{\linewidth}{0.4pt}
\begin{Verbatim}[breaklines=true]
Given a list of actions performed on the website {website_url} and the corresponding screenshots ACTION_LIST: [{action_list}] Your task is to come up with a single task description that will be accomplished by performing these actions in the given sequence on the website.

*IMPORTANT*
0. The task must contain some actions.
1. You should propose tasks that are clear and specific.
2. The task description should provide all the necessary information to complete the task.
3. The task description must indicate the domain of website at the end of the task.
4. The task should be feasible to complete by a real user and should not require any additional information that is not specified in this input.
5. The task description should specify constraints like given budget, product features, and other specifications that can narrow down the search to a particular item/product.
6. If the action sequence uses a visible image as the key cue and refers to the corresponding image URL, preserve that image-conditioned constraint in the summarized task by mentioning the image URL or a distinctive substring/path from it.
7. Do NOT invent or hallucinate image URLs. Only preserve an image URL condition if it is clearly supported by the action sequence.
8. Do NOT use any quotation marks in the task description.
\end{Verbatim}
\end{tcolorbox}
\vspace{-0.5em} %
\captionof{figure}{System prompts used in the Cold-Start data synthesis phase. We explicitly guide the exploring agent to leverage image URLs within the DOM to formulate vision-dependent tasks, followed by a summarization step to finalize the user instruction.}
\label{fig:task_proposal_prompts}
\vspace{1em} %

\subsection{Trajectory-Level Rejection Sampling}
\label{sec:rejection_sampling}

To ensure the quality of our synthesized training data, we employ a multi-stage VLM-based judge pipeline inspired by Online-Mind2Web \cite{xueillusion}\footnote{\url{https://github.com/OSU-NLP-Group/Online-Mind2Web/blob/main/src/methods/webjudge_general_eval.py}}. The evaluation proceeds in three progressive steps: (1) \textbf{Rubric Generation:} The model first generates specific evaluation rubrics based solely on the task instruction. (2) \textbf{State Scoring:} For each step in the trajectory, the VLM assesses the intermediate browser screenshot and assigns a task-progress relevance score according to the generated rubrics. (3) \textbf{Final Judgment:} The state screenshots with a high relevance score, along with the complete sequence of operations, are provided to the VLM to make a final binary (success or failure) decision for the entire trajectory. 

We utilize Gemini-3-Pro \cite{googlepro} across all three stages of this pipeline. When validated on the representative VWA-Lite subset, our automated judge achieves an 83.0\% agreement accuracy compared to the rigorous ground-truth functional evaluators, demonstrating its high reliability for filtering high-quality SFT data without manual intervention.

\section{Agent Action Space \& System Prompts}
\label{sec:action_space_prompts}

In this section, we detail the complete action space and the system prompts used by our multimodal browser-use agent. Table~\ref{tab:action_space} provides a comprehensive definition of all executable actions. The action space is strictly divided into two categories: standard state-changing \textit{Browser Operations} (e.g., page navigation, tab management, and element interaction) and our proposed state-preserving \textit{Probing Primitives}. 

To ensure fair comparison, both the baseline and our P2A agent share the identical baseline prompt for browser operations. As illustrated in Figure~\ref{fig:unified_system_prompt}, our P2A framework augments the baseline prompt with descriptions and usage strategies for the probing primitives, seamlessly integrating "thinking with visualizing" into the agent's operation decision-making without altering the underlying interaction paradigm.

\vspace{1em}
\noindent
\begin{tcolorbox}[
    enhanced,
    breakable,
    colback=gray!5!white, 
    colframe=gray!50!black, 
    title=\textbf{Unified System Prompt for Browser Navigation}, 
    fonttitle=\small\bfseries, 
    arc=2mm, 
    boxrule=0.5pt,
    width=\linewidth,
    left=6mm, right=6mm, top=3mm, bottom=3mm
]
\small
\begin{Verbatim}[breaklines=true, commandchars=\\\{\}]
You are an autonomous intelligent agent tasked with navigating a web browser. You will be given web-based tasks. These tasks will be accomplished through the use of specific actions you can issue.

## Here's the information you'll have:
### The objective: This is the task you're trying to complete.
### The webpage screenshots: These are screenshots of the webpage. Screenshots may show visualized interactable elements (Set-of-Marks (SoM) boxes) drawn by internal observe actions. Each interactable element is assigned a unique numerical id.
{{if w/ probing primitive}}
### Image references: Each image is labeled in the prompt captions as `Image <img_id>` where `<img_id>` is a numeric identifier you can use to refer to that image.
{{/if w/ probing primitive}}

### The text observations: These list the IDs of all interactable elements on the current webpage frame with their text content if any, in the format [id] [tagType] [text content]. `tagType` is the type of the element, such as button, link, or textbox. `text content` is the text content of the element. For example, [1234] [button] ['Add to Cart'] means that there is a button with id 1234 and text content 'Add to Cart' on the current webpage. [] [StaticText] [text] means that the element is of some text that is not interactable.
{{if w/ probing primitive}}
**CRITICAL**: While you have the full list of elements, **text alone is ambiguous**. You lack spatial context (position, occlusion, layout). You MUST use Internal Actions to **visualize** candidate IDs on the screenshot to verify they are the correct target before acting.
### The internal observation history: This records internal observe actions since the last browser action (page step). This history is reset after any browser action.
### The memory notes: These are persistent notes carried across page steps.
{{/if w/ probing primitive}}
### The current webpage's URL: This is the URL of the page you're currently navigating.
### The open tabs: These are the tabs you have open.

### The history of rationales and actions: These are your previous responses and actions taken at each webpage state
{{if w/ probing primitive}}
, along with action-related observations of those turns.
{{/if w/ probing primitive}}
Refer to this history to reason about what you have effectively accomplished so far, so you can decide what to do next.
## Reflections: These are reflections and notes about failed attempts to this task, which can help you determine what to do next.

## The actions you can perform fall into the following categories:
{{if w/ probing primitive}}
   Call actions in the format inside triple backticks (```), using Python-call syntax with literal arguments (e.g. ```click(i=7)```, ```element_focus(ids=["7"], pad="mid")```).
-----------------INTERNAL OBSERVE ACTION----------------
**Usage Strategy**: You act as a detective. Use these tools to **draw bounding boxes (SoM)** on the screenshot. Never rely solely on text descriptions; always **Verify Visually** that an ID corresponds to the correct UI element.
### Internal Observation Actions:
```region_focus(b=[x1,y1,x2,y2], ac=bool, vi=[img_id1,img_id2])```: Inspect a region. Visualizes ALL DOM elements within the bbox to understand layout. Optionally add an image crop of the region to the context (`ac` means `add_crop`).
```element_focus(ids=["id1", "id2"], pad="low"|"mid"|"high", ac=bool, vi=[img_id1,img_id2])```: Inspect elements by DOM IDs and visualize their surrounding region. Pad semantics: `low` = only visualize the provided `ids`; `mid` = visualize ALL elements whose bbox centers fall inside the minimal union bbox of `ids`; `high` = like `mid` but expand that bbox by 1.2x around center. Optionally add an image crop of the region to the context (`ac` means `add_crop`).
```register(b=[x1,y1,x2,y2], l="label of the element", vi=[img_id1,img_id2])```: Register a pseudo element when the real DOM/SoM is missing it. This creates a new numeric id (max_id+1) and adds a line like `[id] [Registered] [label]`. You can then use this id in external actions; the env executes at the bbox center.
```commit(c="comment string", ids=["id"], b=[x1,y1,x2,y2])```: Comment notes and mark findings on elements or region(bbox), which would be saved in memory_notes.
  - `vi`(list(int)) means the images to be visualized(only valid on screenshot and crops) and would not draw anything with an empty list `[]`(default).
  - All `b` arguments are normalized [x1, y1, x2, y2] (0-1000) based on the full screenshot.
-----------------EXTERNAL BROWSER ACTION----------------
{{/if w/ probing primitive}}
### Page Operation Actions:
```click(i=ID)```: This action clicks on an element with a specific id on the webpage.
```type(i=ID, c=CONTENT, e=ENTER_AFTER)```: Use this to type CONTENT into the field with id. If `e` is 0/False, the "Enter" key is not pressed after typing; otherwise, the "Enter" key is automatically pressed.
```hover(i=ID)```: Hover over an element with id.
```press(k=KEY_COMB, c=TEXT)```: Press a keyboard key or key combination (e.g., delete, ctrl+a) with optional text input if the key combination requires it (e.g., ```press(k="ctrl+f", c="some_text")```).
```scroll(d="down")``` or ```scroll(d="up")```: Scroll the webpage up or down.

### Tab Management Actions:
```new_tab()```: Open a new, empty browser tab.
```tab_focus(p=TAB_INDEX)```: Switch the browser's focus to a specific tab using its index.
```close_tab()```: Close the currently active tab.

### URL Navigation Actions:
```goto(u=URL)```: Navigate to a specific URL.
```go_back()```: Navigate to the previously viewed page.
```go_forward()```: Navigate to the next page (if a previous 'go_back' action was performed).

### Completion Action:
```stop(a=ANSWER)```: Issue this action if you believe the task is complete or infeasible. If the objective is to find a text-based answer, provide the answer in the argument. If you deem the task is infeasible, provide a reason why.

## To be successful, it is very important to follow the following rules:
{{if w/ probing primitive}}
1. **Verify-then-Act Protocol**: Do not trust the text observation blindly. Even if you see an ID in the text list, you MUST use **internal action** to **visualize it on the screenshot** first. Only verify the element's position and context visually before issuing an External Action.
2. **Phase Separation**: Use internal actions to build visual evidence until confident. Avoid premature external actions as they invalidate your visual verification.
{{/if w/ probing primitive}}
3. You should only issue one action at a time.
4. Provide your reasoning process wrapped in <thinking>...</thinking> before giving the action.
5. Generate the action in the correct format immediately after the </thinking> tag. For example, "<thinking>reasoning process</thinking>. In summary, the next action I will perform is" phrase, followed by action inside ``````. For example, "<thinking>reasoning process</thinking> In summary, the next action I will perform is ```click(i=1234)```" 
{{if w/ probing primitive}}
or "<thinking>reasoning process</thinking> In summary, the next action I will perform is ```element_focus(ids=[\"1234\"], pad=\"mid\")```".
{{/if w/ probing primitive}}
6. Issue stop action when you think you have achieved the objective. Don't generate anything after stop.

\end{Verbatim}
\end{tcolorbox}
\vspace{-0.5em}
\captionof{figure}{The unified system prompt template. The highlighted conditional blocks (e.g., \texttt{\{\{\#if ...\}\}}) showcase the additional text used by our Probe-to-Act agent. Our framework extends the base capability by simply introducing the tool usage guidelines for visualization probing.}
\label{fig:unified_system_prompt}
\vspace{1em}
\begin{table*}[htbp]
\centering
\caption{The complete action space for the browser-use agent. Action formats explicitly list all required parameters.}
\label{tab:action_space}
\resizebox{\textwidth}{!}{
\small
\begin{tabular}{p{0.31\textwidth} p{0.24\textwidth} p{0.43\textwidth}}
\toprule
\textbf{Action Format} & \textbf{Category} & \textbf{Description} \\
\midrule
\multicolumn{3}{c}{\cellcolor{gray!10}\textbf{Standard Browser Operations}} \\
\midrule
\texttt{click(id)}                                  & Page Operation & Clicks on the specified DOM element. \\
\texttt{type(id, text, enter\_after)}               & Page Operation & Types text into a field, optionally pressing `Enter'. \\
\texttt{hover(id)}                                  & Page Operation & Hovers over the specified DOM element. \\
\texttt{press(key\_comb, text)}                     & Page Operation & Presses a key combination with optional text (e.g., \texttt{ctrl+f}). \\
\texttt{scroll(direction)}                          & Page Operation & Scrolls the webpage \texttt{up} or \texttt{down}. \\
\texttt{new\_tab()}                                 & Tab Management & Opens a new, empty browser tab. \\
\texttt{tab\_focus(tab\_index)}                     & Tab Management & Switches the browser's focus to a specific tab via index. \\
\texttt{close\_tab()}                               & Tab Management & Closes the currently active tab. \\
\texttt{goto(url)}                                  & URL Navigation & Navigates the current tab to a specific URL. \\
\texttt{go\_back()}                                 & URL Navigation & Navigates to the previously viewed page. \\
\texttt{go\_forward()}                              & URL Navigation & Navigates to the next page in history. \\
\texttt{stop(answer)}                               & Completion     & Concludes the task, returning an answer if applicable. \\
\midrule
\multicolumn{3}{c}{\cellcolor{gray!10}\textbf{Probing Primitives (Internal Actions / Ours)}} \\
\midrule
\texttt{region\_focus(bbox, add\_crop)}   & Visualization Probe & Visualizes all DOM elements within the bounding box to clarify local layout. \\
\texttt{element\_focus(ids, pad, add\_crop)} & Visualization Probe & Renders specific target and surrounding elements to verify their exact visual appearance. \\
\texttt{register(bbox, label)}            & Visualization Probe & Assigns a new pseudo-ID to an unparsed, visually salient region, making it actionable. \\
\texttt{commit(comment, ids, bbox)}                           & Memory Probe        & Anchors verified elements or regions with text notes, preserving crucial context in the memory. \\
\bottomrule
\end{tabular}
}
\end{table*}
\section{Implementation \& Training Details}
\label{sec:implementation}

We initialize our model using the \texttt{Qwen3-VL-8B-Instruct} model. During training, we freeze the Vision Transformer (ViT) and perform full-parameter fine-tuning on the remaining components (i.e., the LLM backbone and the vision-language projector). The model is optimized using the AdamW optimizer with a learning rate of $1\times 10^{-5}$ and trained for 1 epoch. Training is conducted on nodes of 8 NVIDIA H800 GPUs utilizing DeepSpeed ZeRO-3 optimization. We set the per-device micro-batch size to 1 and apply gradient accumulation to reach an effective global batch size of 224 per update step. To balance high-resolution perception and computational efficiency, image resolutions are dynamically constrained such that each image is encoded into 16 to 2,500 visual tokens. The maximum context window is strictly truncated at 20,480 tokens to prevent out-of-memory errors during long-horizon rollouts.

At inference, we use greedy decoding ($\tau=0$) and cap each step at 2,048 new tokens. We retry up to 5 times after token-limit, parsing, or invalid-parameter failures. If more than 10 internal probes occur before an operation, a one-time prompt requires an immediate state-changing action to prevent loops. Due to limited website-hosting and deployment resources, all results use a single run. We use the evaluator cleaned by \citet{andrade2025let} on WebArena and VisualWebArena, and the official script with o4-mini~\cite{OpenAI2025o3o4mini} as judge on Online-Mind2Web.

\section{Additional Evaluation Analyses}
\subsection{Complete Inference Accounting}
\label{sec:inference_accounting}
The 30-step budget counts only state-changing browser operations. Probe actions do not consume these outer interaction steps, but are included in probe turns, image inputs, and wall-clock time in Table~\ref{tab:inference_accounting}.
\begin{table*}[htbp]
\centering
\caption{Average per-task inference accounting on VWA. All image inputs and wall-clock measurements include the additional probe turns.}
\label{tab:inference_accounting}
\small
\setlength{\tabcolsep}{8pt}
\renewcommand{\arraystretch}{1.08}
\begin{tabular}{@{}lrrrrr@{}}
\toprule
\textbf{Model} & \textbf{SR} & \textbf{Browser ops} & \textbf{Probe turns} & \textbf{Image inputs} & \textbf{Wall time} \\
\midrule
Qwen3-VL-8B-Instruct & 24.6 & 15.84 & 0 & 16.04 & 203.2 s \\
\quad + P2A (Round 2) & \textbf{32.9} & 11.25 & 5.12 & 16.90 & 243.8 s \\
\quad + P2A (Prompt) & 18.0 & 19.24 & 2.59 & 20.83 & 338.6 s \\
\midrule
Gemini-3-Flash & 42.4 & 12.80 & 0 & 13.20 & 380.1 s \\
\quad + P2A (Prompt) & \textbf{48.7} & 13.08 & 8.14 & 20.70 & 432.8 s \\
\bottomrule
\end{tabular}
\end{table*}
P2A deliberately uses additional test-time computation. For the trained Qwen model, it reduces browser operations from 15.84 to 11.25 while improving success, with only a small increase in image inputs; by contrast, prompt-only P2A uses more image inputs and wall time but performs worse, showing that extra computation alone is insufficient. Gemini exhibits a different profile in which prompt-based probing increases image inputs and improves success with a smaller relative wall-time increase. We therefore report each setting separately and do not claim equal-compute isolation.

\subsection{Robustness to Missing Evidence}
\label{sec:missing_evidence}
Because the standard benchmarks do not isolate off-DOM targets or annotate cross-page dependencies, we construct controlled stress tests by randomly removing current-page DOM records or prior history turns while preserving the underlying browser state.
\begin{table*}[htbp]
\centering
\caption{Controlled robustness to missing evidence on VWA. (a) Interactable DOM records and their SoM marks are removed while screenshot pixels and true interactivity are preserved. (b) Complete prior turns are removed while the current screenshot, DOM, and browser state are preserved.}
\label{tab:missing_evidence_robustness}
\scriptsize
\setlength{\tabcolsep}{4pt}
\renewcommand{\arraystretch}{1.08}
\textbf{(a) Missing DOM evidence.}\\[0.3em]
\resizebox{\textwidth}{!}{%
\begin{tabular}{@{}rrrrrrrr@{}}
\toprule
\textbf{Removed DOM} & \textbf{Standard DOM+SoM SR} & \textbf{$\Delta$ clean} & \textbf{P2A SR} & \textbf{$\Delta$ clean} & \textbf{P2A w/o Registration SR} & \textbf{$\Delta$ clean} & \textbf{Registration gain} \\
\midrule
0\%  & 24.6 & --   & 32.9 & --   & 32.3 & --    & +0.6 \\
10\% & 22.4 & -2.2 & 31.0 & -1.9 & 26.4 & -5.9  & +4.6 \\
30\% & 16.0 & -8.6 & 27.1 & -5.8 & 22.3 & -10.0 & +4.8 \\
\bottomrule
\end{tabular}%
}

\vspace{0.7em}
\textbf{(b) Missing history evidence.}\\[0.3em]
\resizebox{\textwidth}{!}{%
\begin{tabular}{@{}rrrrrrrr@{}}
\toprule
\textbf{Removed history} & \textbf{Standard DOM+SoM SR} & \textbf{$\Delta$ clean} & \textbf{P2A SR} & \textbf{$\Delta$ clean} & \textbf{P2A w/o Commit SR} & \textbf{$\Delta$ clean} & \textbf{Commit gain} \\
\midrule
0\%  & 24.6 & --   & 32.9 & --   & 31.5 & --   & +1.4 \\
30\% & 19.2 & -5.4 & 28.0 & -4.9 & 25.7 & -5.8 & +2.3 \\
50\% & 18.0 & -6.6 & 26.3 & -6.6 & 24.3 & -7.2 & +2.0 \\
\bottomrule
\end{tabular}%
}
\end{table*}
Full P2A degrades more gracefully than the corresponding primitive ablations. Registration provides only a small gain on clean inputs but a stable 4.6--4.8 point gain once structured records are missing, while Commit retains a 2.0--2.3 point advantage when prior turns are removed. The standard DOM+SoM agent also falls below full P2A under both stress tests. These controlled tests establish robustness to missing evidence rather than estimating how often such corruption occurs naturally.

\section{Qualitative Case Studies}
\label{sec:qualitative examples}
\begin{figure*}[p]
    \centering
    \includegraphics[width=\textwidth]{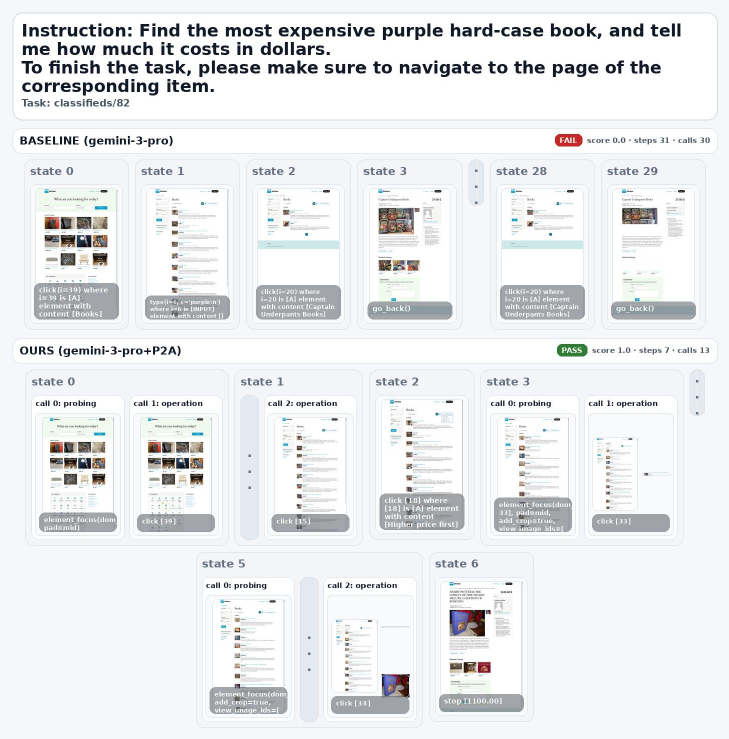}
    \caption{\textbf{Comparison of Gemini-3-Pro with and without P2A.} The task requires finding an item with specific visual features on a shopping site. Cluttered by global Set-of-Marks (SoM) overlays, the baseline struggles to perceive small product thumbnails, leading to a failed trial-and-error strategy of repeatedly opening and closing detail pages. In contrast, P2A identifies visual cues on uncluttered screenshots and utilizes the \textbf{\texttt{focus} primitive} to zoom in and verify the target, completing the task efficiently without any backtracking.}
    \label{fig:gemini_classifieds_82}
\end{figure*}

\begin{figure*}[p]
    \centering
    \includegraphics[width=0.7\textwidth]{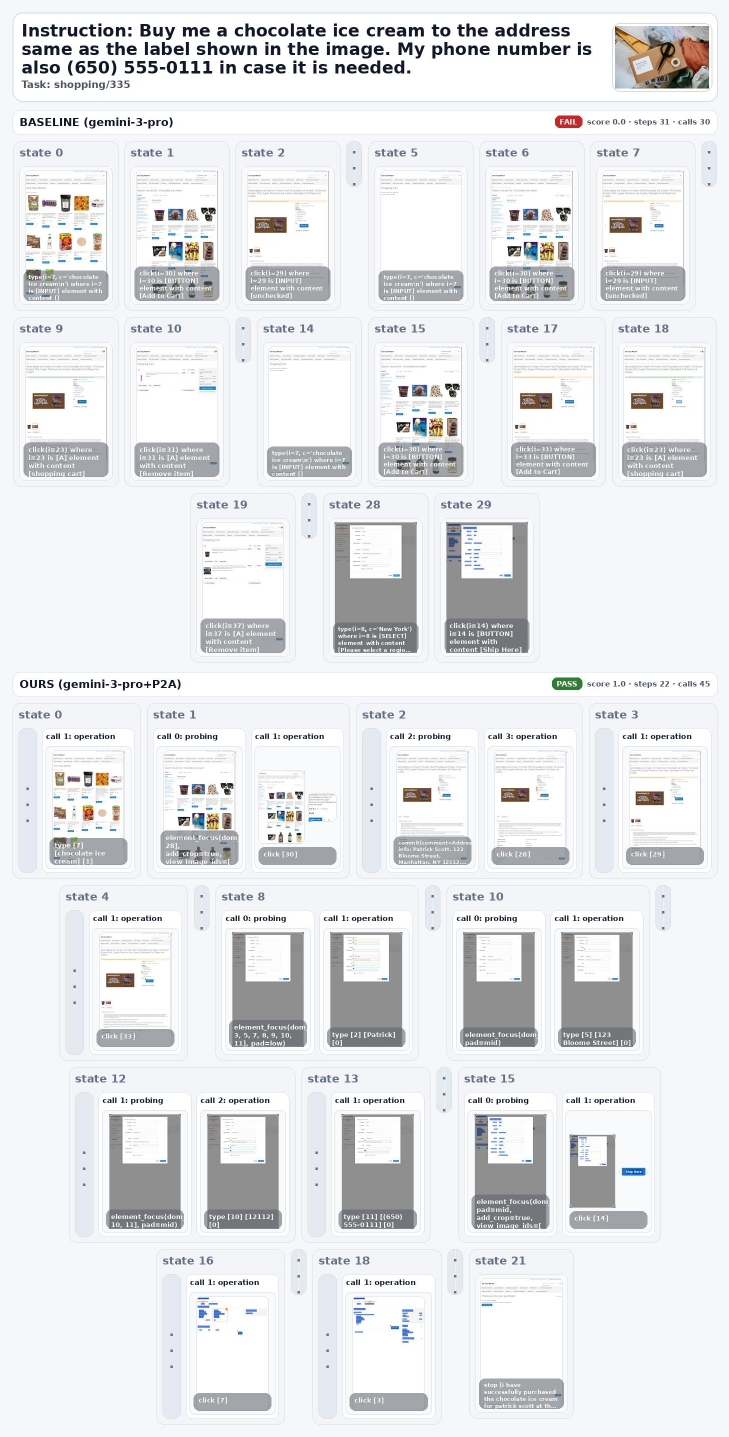}
    \caption{\textbf{Gemini-3-Pro execution with and without P2A.} The task involves finding an item and filling a form using a reference image. Relying solely on action history, the baseline loses visual context and falls into a repetitive purchasing loop (States 5--19). In contrast, P2A uses an evidence-based memory to track progress. By deploying the \textbf{\texttt{commit} primitive} (State 2, Call 2) to proactively save reference details, P2A decouples information extraction from form execution, effectively mitigating errors caused by simultaneous multi-image understanding.}
    \label{fig:gemini_shopping_335}
\end{figure*}

\begin{figure*}[p]
    \centering
    \includegraphics[width=\textwidth]{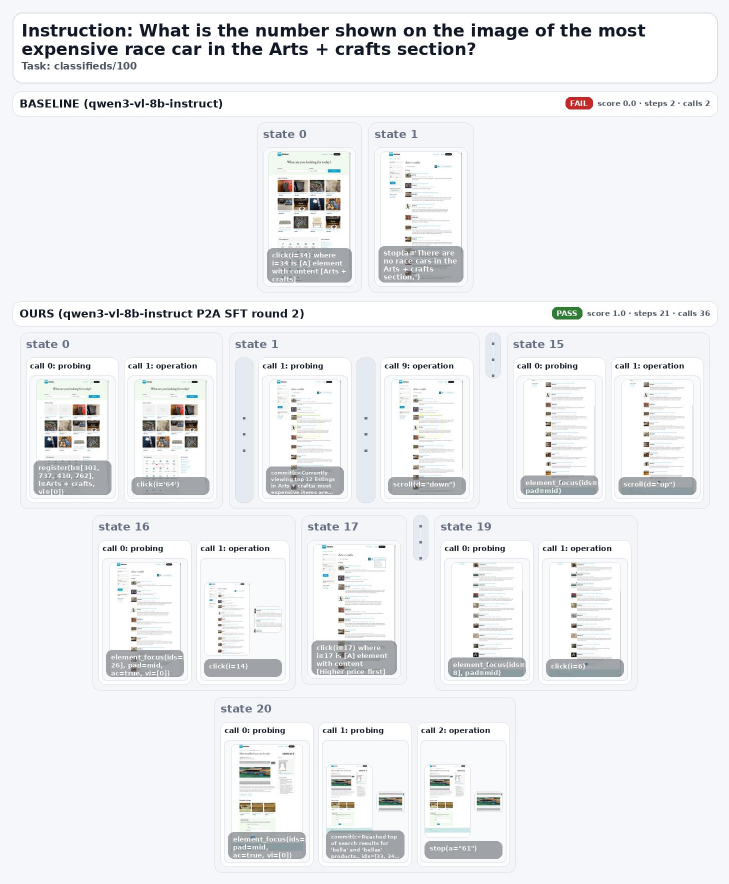}
    \caption{\textbf{Comparison of the baseline Qwen3-VL-8B and our P2A-tuned model.} Without any web exploration, the baseline prematurely halts, falsely concluding the task is unachievable. In contrast, our agent employs the \textbf{\texttt{commit}} primitive (State 1, Call 1) to explicitly record its initial unsuccessful attempt within the task-specified section. Guided by this retained memory, the agent shifts its exploration strategy by sorting products from highest to lowest price (State 17). This efficient search successfully locates the target page, where the agent then utilizes the \textbf{\texttt{focus}} primitive (State 20, Call 0) to zoom in and accurately extract the required fine-grained visual features.}
    \label{fig:qwen_classifieds_100}
\end{figure*}

\begin{figure*}[p]
    \centering
    \includegraphics[width=\textwidth]{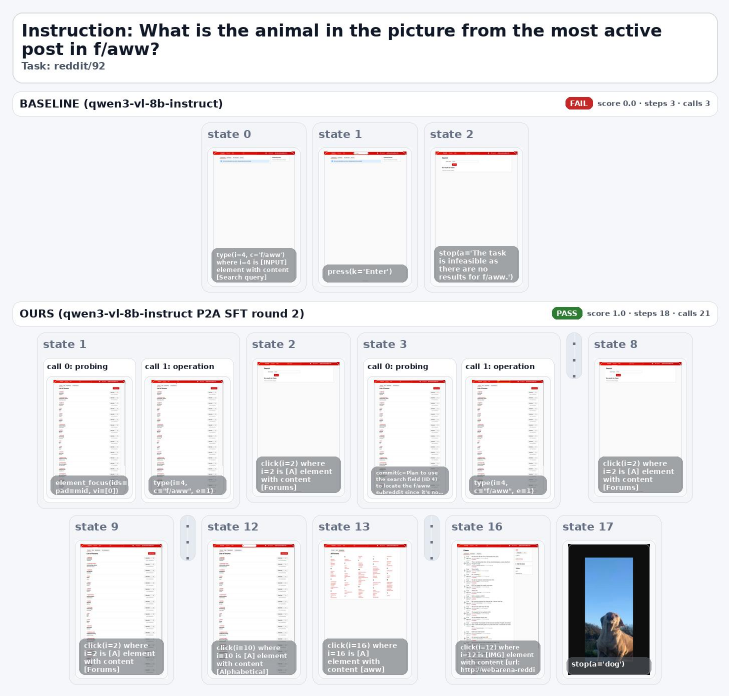}
    \caption{\textbf{Comparison of the baseline Qwen3-VL-8B and our P2A-tuned model.} The baseline prematurely halts without exploring the webpage, falsely concluding that the task is unachievable. Conversely, our agent uses the \textbf{\texttt{commit}} primitive (State 3, Call 0) to explicitly document its failed initial attempt of directly searching for the target forum (``f/aww''). Acknowledging this failure through its memory, the agent strategically adapts by sorting the forums alphabetically (State 12). This efficient workaround successfully locates the correct page, enabling the agent to extract the requested visual information.}
    \label{fig:qwen_reddit_92}
\end{figure*}

\begin{figure*}[p]
    \centering
    \includegraphics[width=\textwidth]{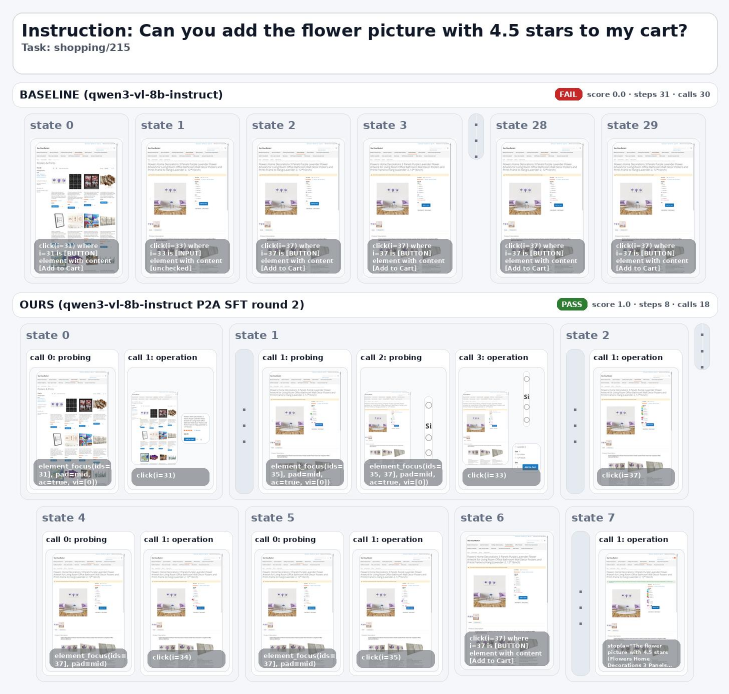}
    \caption{\textbf{Comparison of the baseline Qwen3-VL-8B and our P2A-tuned model.} The task requires adding a product with specific visual features to the cart. Although the baseline locates the correct item, it fails to perceive critical visual feedback (a pop-up prompting to select all specifications first) and falls into a futile loop of repeatedly clicking the ``Add to Cart'' button (States 1--29). In contrast, our agent accurately interprets the dynamic feedback pop-up, selects the required product variants, and successfully completes the addition.}
    \label{fig:qwen_shopping_215}
\end{figure*}

In this section, we provide several qualitative case studies (see Figures~\ref{fig:gemini_classifieds_82}, \ref{fig:gemini_shopping_335}, \ref{fig:qwen_classifieds_100}, \ref{fig:qwen_reddit_92}, and \ref{fig:qwen_shopping_215}) that compare the execution trajectories of our Probe-to-Act (P2A) framework against standard baseline methods on identical tasks. These examples intuitively demonstrate how our probing primitives play a crucial role in: (1) accurately \textbf{grounding} complex task instructions to dense webpage content (Figures~\ref{fig:gemini_classifieds_82}, \ref{fig:qwen_classifieds_100}), (2) robustly \textbf{tracking} long-horizon task progress (Figure~\ref{fig:gemini_shopping_335}), (3) explicitly \textbf{retaining memory} of previous failed attempts to avoid repetitive errors (Figures~\ref{fig:qwen_classifieds_100}, \ref{fig:qwen_reddit_92}), and (4) \textbf{perceiving fine-grained feedback} from the dynamic browser environment (Figure~\ref{fig:qwen_shopping_215}). Therefore, P2A fundamentally prevents premature or erroneous actions, directly contributing to significant improvements in the overall task success rate. Detailed step-by-step analyses are provided in the respective figure captions.
\section{Statements on Research Practices}
\label{app:artifact-ai-use}
\paragraph{Scientific artifacts.}
This work uses existing browser-agent benchmarks and environments, including VisualWebArena, WebArena, and Online-Mind2Web, following their released evaluation protocols. We also use the Qwen3-VL-8B checkpoint as the open-weight base model and incorporate GUI grounding data from ScaleCUA during training. In addition, we create data-synthesis and evaluation scripts for studying active visual probing in browser-use agents. We will open-source the evaluation code and trajectory-synthesis code upon acceptance.
\paragraph{Licenses and terms.}
We use third-party benchmarks, models, and datasets under their respective licenses or terms of use and will cite these original sources. Any artifacts released from this work are intended for research use and will therefore follow the original licenses and usage policies of the corresponding resources.
\paragraph{Use of AI assistants.}
AI assistants were used to support literature search, early-stage idea discussion, and rough sketches for figures before manual refinement by the authors. 
\paragraph{Potential risks.}
The broader area of browser and GUI agents may raise risks in real-world deployment, such as unintended actions, privacy leakage, or misuse for harmful automation. Our experiments are mainly limited to controlled benchmark environments, and practical systems should include safeguards such as sandboxing, action auditing, and human confirmation for sensitive operations.
\clearpage %

\end{document}